\documentclass[lettersize,journal]{IEEEtran}
\usepackage{amsmath,amsfonts,amssymb}
\usepackage{algorithmic}
\usepackage{algorithm}
\usepackage{array}
\usepackage[caption=false,font=normalsize,labelfont=sf,textfont=sf]{subfig}
\usepackage{textcomp}
\usepackage{stfloats}
\usepackage{url}
\usepackage{verbatim}
\usepackage{graphicx}
\usepackage{cite}

\usepackage{multirow}
\usepackage{colortbl}
\usepackage{xcolor}
\usepackage{color}
\usepackage{bbm}
\usepackage{xspace}
\usepackage{graphicx}
\usepackage{float}
\usepackage{makecell}
\usepackage{tcolorbox}
\usepackage{tikz}
\usepackage{array}
\usepackage{caption}
\usepackage{subcaption}

\def\eg{\textit{e.g.}}
\def\ie{\textit{i.e.}}

\definecolor{blue}{rgb}{0, 0, 0}

\begin{document}

\title{DeepInteraction++: Multi-Modality Interaction for Autonomous Driving}

\author{Zeyu Yang$^{\ast}$, Nan Song$^{\ast}$, Wei Li$^{\ast}$, Xiatian Zhu, Li Zhang, Philip H.S. Torr
\thanks{$^{\ast}$ Equal contribution.}
\thanks{
Zeyu Yang, Nan Song, and Li Zhang are with the School of Data Science, at Fudan University. 
Wei Li is with Nanyang Technological University. 
Xiatian Zhu is with the University of Surrey.
Philip H.S. Torr is with the University of Oxford.}

}

\markboth{Journal of \LaTeX\ Class Files,~Vol.~14, No.~8, August~2021}
{Shell \MakeLowercase{\textit{et al.}}: A Sample Article Using IEEEtran.cls for IEEE Journals}

\maketitle

\begin{abstract}

Existing top-performance autonomous driving systems typically rely on the {\em multi-modal fusion} strategy for reliable scene understanding. This design is however fundamentally restricted due to overlooking the modality-specific strengths and finally hampering the model performance. To address this limitation, in this work, we introduce a novel {\em modality interaction} strategy that allows individual per-modality representations to be learned and maintained throughout, enabling their unique characteristics to be exploited during the whole perception pipeline.
To demonstrate the effectiveness of the proposed strategy, we design {\em DeepInteraction++}, a multi-modal interaction framework characterized by a multi-modal representational interaction encoder and a multi-modal predictive interaction decoder. 
Specifically, the encoder is implemented as a dual-stream Transformer with specialized attention operation for information exchange and integration between separate modality-specific representations. Our multi-modal representational learning incorporates both object-centric, precise sampling-based feature alignment and global dense information spreading, essential for the more challenging planning task.
The decoder is designed to iteratively refine the predictions by alternately aggregating information from separate representations in a unified modality-agnostic manner, realizing multi-modal predictive interaction.
Extensive experiments demonstrate the superior performance of the proposed framework on both 3D object detection and end-to-end autonomous driving tasks. Our code is available at https://github.com/fudan-zvg/DeepInteraction.

\end{abstract}


\section{Introduction}

Safe autonomous driving relies on reliable scene perception, with 3D object detection as a core task by localizing and recognizing decision-sensitive objects in the surrounding 3D world. For stronger perception capability, LiDAR and camera sensors have been simultaneously deployed in most current autonomous vehicles to provide point clouds and RGB images respectively.
The two modalities exhibit naturally strong complementary effects
due to their different perceiving characteristics.
Point clouds involve necessary localization and geometry information with sparse representation, while images offer rich appearance and semantic information at high resolution.
Therefore, dedicated {\em information fusion} across modalities becomes particularly crucial for strong scene perception.

\begin{figure}
    \centering
    \includegraphics[width=1.0\linewidth]{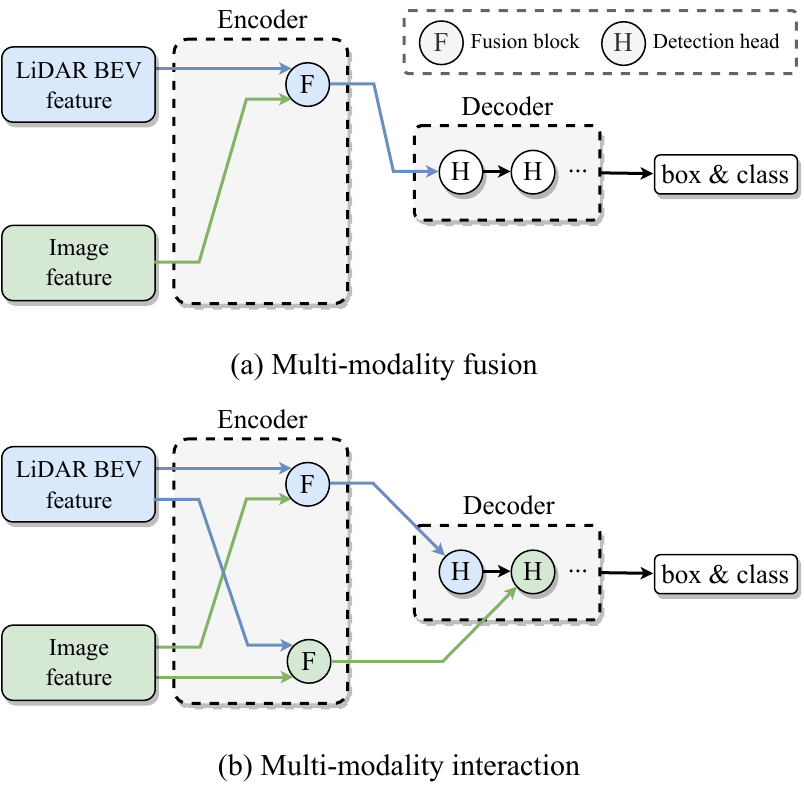}
    \caption{
    \textbf{Schematic strategy comparison}. 
    \textbf{(a)} 
    Existing multi-modality fusion-based 3D detection: Fusing individual per-modality representations into a single hybrid representation from which the detection results are further decoded.
    \textbf{(b)} 
    Our multi-modality interaction-based 3D detection: Maintaining two modality-specific representations throughout the whole pipeline with
    both \textit{\bf \em representational interaction} in the encoder and
    \textit{\bf \em predictive interaction} in the decoder.
}
    \label{fig:teaser}
    \vspace{-5mm}
\end{figure}

Taking the quintessential and pivotal perception task of 3D object detection as an example, existing multi-modal 3D objection detection methods typically
adopt a {\bf\em modality fusion} strategy (Figure~\ref{fig:teaser}(a)) by combining individual per-modality representations into a {\em single} hybrid representation.
For instance, PointPainting~\cite{vora2020pointpainting} and its variants~\cite{wang2021pointaugmenting,Yin2021MVP,Xu2021FusionPaintingMF} aggregate category scores or semantic features from the image space into the 3D point cloud space. AutoAlign~\cite{chen2022autoalign} and VFF~\cite{li2022vff} similarly integrate image representations into the 3D grid space. Latest alternatives~\cite{li2022deepfusion,liu2022bevfusion,liang2022bevfusion} merge the image and point cloud features into a joint bird's-eye view (BEV) representation. 
This fusion approach is, however, structurally restricted due to its intrinsic limitation of potentially dropping off a large fraction of modality-specific representational strengths due to largely imperfect information fusion into a unified representation.

To overcome the aforementioned limitations, in this work 
a novel {\bf\em modality interaction} strategy, termed {\bf DeepInteraction++},  for integrating information from different sensors is introduced (Figure~\ref{fig:teaser}(b)).
Our key idea is to learn and maintain multiple modality-specific representations instead of deriving a single fused representation. This approach enables inter-modality interaction, allowing for the spontaneous exchange of information and the retention of modality-specific strengths with minimal interference between them.
Specifically, we start by mapping 3D point clouds and 2D multi-view images into the multi-scale LiDAR BEV features and perspective camera features using two separate feature backbones in parallel. 
Subsequently, with an encoder we interact heterogeneous features for progressive representation learning and integration in a {\em bilateral} manner. To fully exploit per-modality representations, we design a decoder to conduct iteratively multi-modal predictive interaction to yield more accurate perception results.

Our {\bf contributions} can be summarized as follows:
{\bf (i)} 
We introduce a novel {\em modality interaction} strategy for multi-modal learning for autonomous driving tasks,
addressing a fundamental limitation of the previous {\em modality fusion} strategy in exploiting the modality-specific information.
{\bf (ii)}
We formulate the DeepInteraction++ architecture, characterized by a multi-modal predictive interaction decoder and a multi-modal representational interaction encoder, leveraging a powerful dual-stream Transformer architecture and meticulously curated interaction operations.
{\bf (iii)} 
Extensive experiments on the highly competitive nuScenes dataset demonstrate the superiority of our methods over prior art models.
Beyond the 3D object detection, we also evaluate the proposed framework on end-to-end autonomous driving to demonstrate the efficacy of the proposed {\em modality interaction} philosophy more thoroughly, benefiting from the flexible multi-modal interaction design, 
In particular, DeepInteration++ not only effectively extracts object-centric information to achieve strong 3D object detection capabilities, but is also capable of constructing dense representations of the surrounding environment, offering a versatile solution for various autonomous driving tasks.

A preliminary version of this work (DeepInteraction \cite{yang2022deepinteraction}) was presented as spotlight at NeurIPS 2022. 
In this extended paper, 
we further upgrade the proposed paradigm of {\em multi-modality interaction}
in both module design and architecture expansion.
{\bf (1)} 
We equip {the encoder} with a dual-stream Transformer architecture for integrating intra-modal representational learning and inter-modal representational learning simultaneously.
Compared with the original 
FFN-based representation integration,
this new design offers higher scalability and computational overhead reduction.
{\bf (2)} 
We replace the stand-alone attention originally used for intra-modal interactions with deformable attention, enabling a more flexible receptive field and multi-scale interactions.
{\bf (3)} 
We additionally introduce LiDAR-guided cross-plane polar ray attention for propagating the underlying semantics from the visual representation 
to the LiDAR representation in a dense manner. 
This is achieved by learning the inherent correspondence between the BEV polar ray and the camera imaging column.
The motivation is to provide a rich dense context to complement the original object-centric sparse interaction.
{\bf (4)}
To further improve the runtime and memory demands, 
we introduce grouped sparse attention, without compromising performance, and creating extra room for further scaling our approach.
{\bf (5)} 
We expand the applications of the approach from 3D object detection as originally focused on, to more diverse autonomous driving tasks (\eg, end-to-end prediction and planning).
This is made possible due to our more efficient and capable multi-modal learning architecture design.
Exploring this multi-task strategy in a single architecture not only demonstrates
the generic applicability and scalability of our approach but also suggests a feasible strategy of designing autonomous driving system in practice. 
{\bf (6)} 
We evaluate and compare the latest detection methods, showing that our interaction-focused multi-modal representation learning framework is superior in comparison.
{\bf (7)} We conduct more extensive ablation experiments ranging from parameter choices to module designs, elucidating the sources of performance enhancement and systematically exploring the scalability of our framework.

\section{Related work}

\paragraph{3D object detection with single modality}
Although automated driving vehicles are generally equipped with both LiDAR and multiple surround-view cameras, many previous methods still focus on resolving 3D object detection by exploiting data captured from only a single form of sensor. For camera-based 3D object detection, since depth information is not directly accessible from RGB images, some previous works~\cite{huang2021bevdet,wang2019pseudo,reading2021caddn} lift 2D features into a 3D space by conducting depth estimation, followed by performing object detection in the 3D space. Another line of works~\cite{wang2022detr3d,liu2022petr,li2022bevformer,lu2022ego3rt,jiang2022polar, misra2021-3detr, wang2023exploring, liu2023sparsebev} resort to the detection Transformer~\cite{carion2020detr} architecture. They leverage 3D object queries and 3D-2D correspondence to incorporate 3D computation into the detection pipelines.

Despite the rapid progress of camera-based approaches, the state-of-the-art of 3D object detection is still dominated by LiDAR-based methods. Most of the LiDAR-based detectors quantify point clouds into regular grid structures such as voxels~\cite{zhou2018voxelnet,yan2018second}, pillars~\cite{pointpillars, li2023pillarnext} or range images~\cite{bewley2020range,fan2021rangedet,Chai2021ToTP} before processing them. Due to the sampling characteristics of LiDAR, these grids are naturally sparse and hence fit the Transformer design. So a number of approaches~\cite{mao2021votr,fan2021sst} have applied the Transformer for point cloud feature extraction. Differently, several methods use the Transformer decoder or its variants as their detection head~\cite{bai2022transfusion, wang2021objectdgcnn}. 
Due to intrinsic limitations with either sensor,
these methods are largely limited in performance.

\vspace{0.5em}

\paragraph{Multi-modality fusion for 3D object detection}
Leveraging the perception data from both camera and LiDAR sensors
usually provides a more sound solution and leads to better performance.
This approach has emerged as a promising direction.
Existing 3D detection methods typically perform multi-modal fusion
at one of the three stages: raw input, intermediate feature, and object proposal.
For example, PointPainting~\cite{vora2020pointpainting} is the pioneering input fusion method~\cite{wang2021pointaugmenting,huang2020epnet, Yin2021MVP}. 
The main idea is to decorate the 3D point clouds with the category scores or semantic features from the 2D instance segmentation network.

Whilst 4D-Net~\cite{piergiovanni20214d} placed the fusion module in the point cloud feature extractor to allow the point cloud features to dynamically attend to the image features. ImVoteNet~\cite{qi2020imvotenet} injects visual information into a set of 3D seed points abstracted from raw point clouds. 

The proposal-based fusion methods~\cite{ku2018joint,chen2022futr3d} keep the feature extraction of two modalities independently and aggregate multi-modal features via proposals or queries at the detection head. 
The first two categories of methods take a unilateral fusion strategy 
with a bias to 3D LiDAR modality due to the superiority of point clouds in distance and spatial perception.
Instead, the last category fully ignores the intrinsic association between the two modalities in representation.
As a result, all the above methods fail to fully exploit both modalities, in particular their strong complementary nature.

Besides, a couple of works have explored the fusion of the two modalities in a shared representation space~\cite{liu2022bevfusion,liang2022bevfusion, huang2023detecting, wang2023unitr}. 
They conduct view transformation in the same way~\cite{philion2020lift} as in the camera-only approach. This design is however less effective in exploiting
the spatial cues of point clouds during view transformation,
potentially compromising the quality of camera BEV representation.
This gives rise to an extra need for calibrating such misalignment in network capacity.
To address the efficiency problem, recent methods~\cite{li2024sparsefusion, xie2023sparsefusion} introduce a sparse mechanism to process modality features or object queries, while still restricted in a single fusion manner.

In this work, we address the aforementioned limitations in all previous solutions with a novel multi-modal interaction strategy. 
The key insight behind our approach is that we maintain two modality-specific feature representations and conduct {\em representational} and {\em predictive} interactions for 
maximally exploring their complementary benefits whilst
preserving their respective strengths.

\vspace{0.5em}

\paragraph{End-to-end autonomous driving pipeline.}
Instead of focusing on the perception tasks in the field of autonomous driving, recent approaches~\cite{casas2021mp3, hu2022st, hu2023planning, jiang2023vad} are delving into the end-to-end framework that can simultaneously execute joint tasks from scene perception to ego-planning. Benefiting from explicit and interpretable intermediate results, these methods realize a remarkable breakthrough in the planning task. However, they are still limited to single input modality (especially camera) and perception mode (e.g. BEV or surround view), hindering further improvement. By involving the distinct fusion perception modes of LiDAR and camera input, in contrast, the end-to-end extension of DeepInteraction++ can achieve better performance across various evaluation metrics. Similarly, CamLiFlow~\cite{liu2022camliflow, liu2023learning} also demonstrated the feasibility of applying the bidirectional fusion paradigm to other tasks by successfully adopting this paradigm in the joint estimation of optical flow and scene flow.

\section{DeepInteraction++: 3D object detection via modality interaction}
\label{sec:method}

Most existing 3D object detection frameworks merge data or features from different modalities at specific stages for subsequent feature extraction and decoding. 
At the presence of distinct nature and optimization dynamics of representations from heterogeneous modalities, such an {\em unilateral fusion} may impair detection performance, regardless of whether this integration occurs at an early or late stage in the detection pipeline.
In general, early fusion might restrict the full exploitation of each modality's unique representational learning capabilities, whereas fusion at a later stage can diminish the advantages offered by multi-modal information.
In this paper, we advocate for the modality interaction approach in multi-modal representation learning, allowing mutual enhancement between multi-modal representations while fully leveraging the unique feature extraction advantages of each modality.

Specifically, we propose a novel framework, DeepInteraction++.
In contrast to prior arts, 
it maintains two distinct representations for LiDAR point cloud and camera image modalities throughout the entire detection pipeline while
achieving information exchange and aggregation via multi-modal interaction, instead of creating a single fused representation. 
As shown in Figure~\ref{fig:teaser}(b), it consists of two main components: an encoder with multi-modal representational
interaction (Section \ref{sec:encoder}), and a decoder with multi-modal predictive interaction (Section \ref{sec:decoder}). The encoder realizes information exchange and integration between modalities while maintaining individual per-modality scene representations via multi-modal representational interaction.
The decoder aggregates information from separate modality-specific representations and iteratively refines detection results in a unified modality-agnostic manner, \ie, multi-modal predictive interaction.

\subsection{Encoder: Multi-modal representational interaction}
\label{sec:encoder}
\begin{figure*}
    \centering

    \begin{tabular}{cc}
    \multicolumn{2}{c}{
    \includegraphics[width=0.97\linewidth]{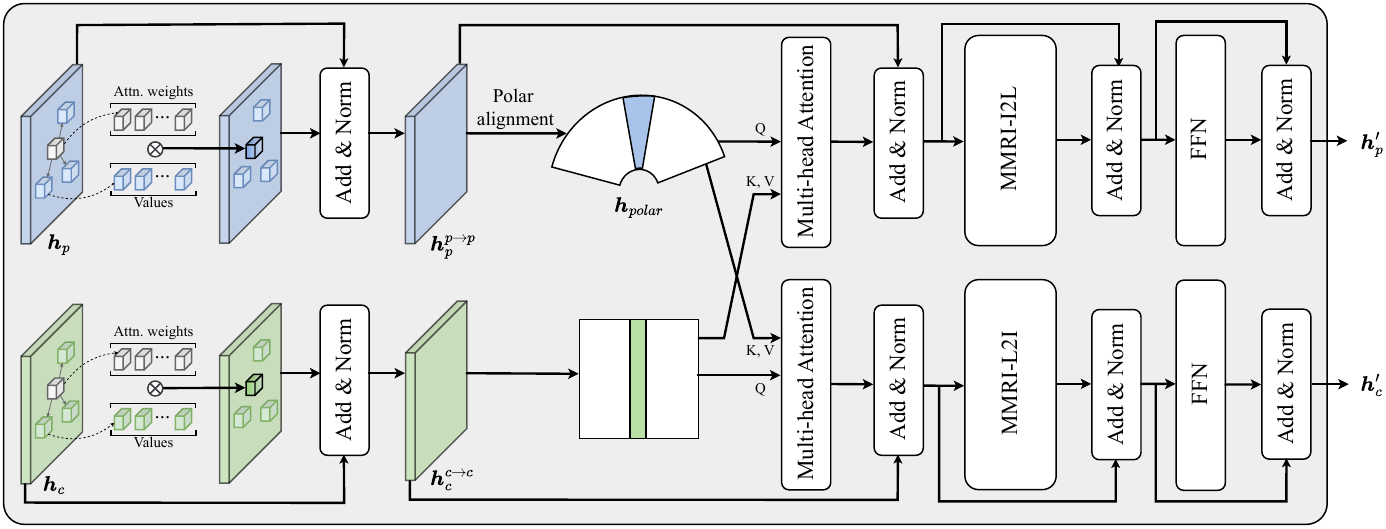}
    } 
 \\
 \multicolumn{2}{c}{
    (a) Overview of the multi-modal representational interaction encoder layer.
 } 
 \\
\includegraphics[width=0.47\linewidth]{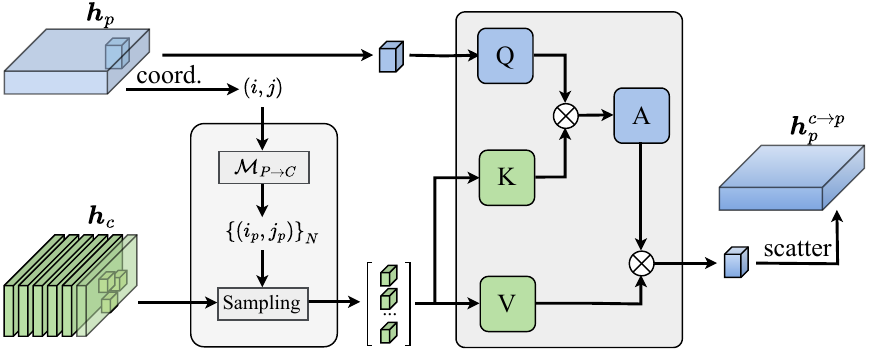} \vline & \includegraphics[width=0.47\linewidth]{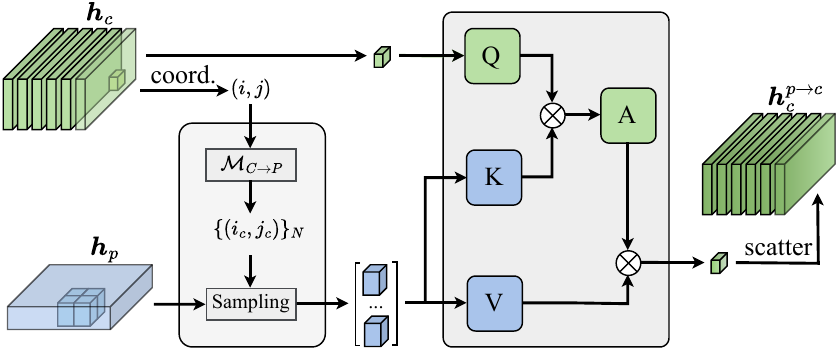}
 \\
(b) {\small MMRI-I2L} & (c) {\small MMRI-L2I}
    \end{tabular}
    
    \caption{
    Structure of our multi-modal representational interaction encoder. \textbf{(a)} Overall architecture: Given two modality-specific representations, the image-to-LiDAR feature interaction \textbf{(b)} spreads the visual signal in the image representation to the LiDAR BEV representation, and the LiDAR-to-image feature interaction \textbf{(c)} takes cross-modal relative contexts from LiDAR representation to enhance the image representations.
    }
    \label{fig:encoder}
    \vspace{-5mm}
\end{figure*}

Unlike conventional modality fusion strategy that often aggregates multi-modal inputs into a hybrid feature map, individual per-modality representations are maintained and enhanced via 
\textit{multi-modal representational interaction} within our encoder.
The encoder is formulated as a {\em multi-input-multi-output} (MIMO) structure, as depicted in Figure~\ref{fig:encoder}(a). It takes two modality-specific scene representations independently extracted by the LiDAR and image backbone as inputs and produces two refined representations as outputs.
Specifically, it is composed by stacking several multi-modal representational interaction encoder layers.
Within each layer, features from different modalities engage in multi-modal representational interaction (MMRI) and intra-modal representational learning (IML), for the inter-modal and intra-modal interactions. We will now outline the overall structure of the encoder.

\noindent \textit{A.1 Interaction encoder with a dual-stream Transformer}

The representational integration approach employed in our preliminary model, DeepInteraction~\cite{yang2022deepinteraction}, has achieved strong results, for enhanced extensibility and usability. In this work, we further push higher scalability and computational overhead reduction.
This is realized by replacing the original encoder layer with a pair of Transformer layers equipped with the customized attention interacting mechanism.
Additionally, the parallel intra-modal and inter-modal representational learning in the original MMRI block are now used as self-attention and cross-attention operations in the refactor architecture. 

Taking the LiDAR branch as an example, the computation within each Transformer layer can be formulated as:
\begin{equation}
\begin{aligned}
\label{eq:tranformer}
    \boldsymbol{h}_{p}^{p\rightarrow p} &=  \mathrm{LN} \left( \mathrm{SA} \left( \boldsymbol{h}_{p} \right) + \boldsymbol{h}_{p} \right), \\
    \boldsymbol{h}_{p}^{c\rightarrow p} &=  \mathrm{LN} \left( \mathrm{CA} \left( \boldsymbol{h}_{p}^{p\rightarrow p}, \boldsymbol{h}_{c} \right) + \boldsymbol{h}_{p}^{p\rightarrow p} \right), \\
    \boldsymbol{h}_{p}^{\prime} &=  \mathrm{LN} \left( \mathrm{FFN} \left( \boldsymbol{h}_{p}^{c\rightarrow p} \right) + \boldsymbol{h}_{p}^{c\rightarrow p} \right), \\
\end{aligned}
\end{equation}
where the $\mathrm{FFN}$ denotes the feed-forward network, $\mathrm{LN}$ denotes Layer Normalization~\cite{vaswani2017attention}, $\mathrm{SA}$ and $\mathrm{CA}$ are instantiated as the MMRI and the IML, respectively. 
The Transformer layer within the image branch follows a similar design. Subsequently, we will detail the computations in each module.

\noindent \textit{A.2 Multi-modal representational interaction (MMRI)}

Taking the representations of two modalities, \ie, the camera panoramic representation $\boldsymbol{h}_c$ and the LiDAR BEV representation $\boldsymbol{h}_p$, as inputs, our multi-modal representational interaction aims to exchange the {\em neighboring context} in a bilateral manner. 

\textbf{Cross-modal correspondence mapping and sampling.} 
To define cross-modality adjacency, we first need to build the pixel-to-pixel(s) correspondence between the representations $\boldsymbol{h}_p$ and $\boldsymbol{h}_c$.
To that end, we construct dense mappings between the image coordinate system $c$ and the BEV coordinate system $p$ ($\mathcal{M}_{p\rightarrow c}$ and $\mathcal{M}_{c\rightarrow p}$).

{\em From Camera image to LiDAR BEV coordinate} $\mathcal{M}_{c\rightarrow p}:\mathbb{R}^2 \rightarrow 2^{\mathbb{R}^2}$
(Figure~\ref{fig:encoder}(c)):
We first project each point $(x,y,z)$ in a 3D point cloud to multi-camera images to form a sparse depth map $\boldsymbol{d}_{sparse}$, followed by depth completion~\cite{ku2018defense} leading to a dense depth map $\boldsymbol{d}_{dense}$.
We further utilize $\boldsymbol{d}_{dense}$ to lift each pixel in the image space into the 3D world space. This results in the corresponding 3D coordinate $(x,y,z)$ given an image pixel $(i,j)$ with depth $\boldsymbol{d}_{dense}^{[i,j]}$. Next, $(x,y)$ is used to locate the corresponding BEV coordinate $(i_p,j_p)=\left( \frac{y-y_{\mathrm{min}}}{y_{\mathrm{max}}-y_{\mathrm{min}}} \times H, \frac{x-x_{\mathrm{min}}}{x_{\mathrm{max}}-x_{\mathrm{min}}} \times W \right)$, where $\left( 
x_{\mathrm{min}},y_{\mathrm{min}},x_{\mathrm{max}},y_{\mathrm{max}} \right)$ is the detection range, and $(H,W)$ is the size of $\boldsymbol{h}_p$.
Denote the above mapping as $T(i,j)=(i_p,j_p)$, 
we can obtain the cross-modal neighbors from the camera to LIDAR BEV via $(2k+1) \times (2k+1)$ sized grid sampling as
$\mathcal{M}_{c\rightarrow p}(i,j) \triangleq \left\{ T(i+\Delta i, j+\Delta j) | \Delta i , ~\Delta j \in [-k, +k] \right\} $.

{\em From LiDAR BEV to Camera image coordinate} $\mathcal{M}_{p\rightarrow c}:\mathbb{R}^2 \rightarrow 2^{\mathbb{R}^2}$ (Figure~\ref{fig:encoder}(b)):
Given a coordinate $(i_p,j_p)$ in BEV, we first obtain the $N$ LiDAR points $P=\left \{(x,y,z)_{n}\right \}_{n=1}^{N}$ within the pillar corresponding to $(i_p,j_p)$.
Then we project these 3D points into camera image coordinate frame $P_{c} = \left\{ (i,j) | (i,j) = \mathrm{Proj} \left((x,y,z),E,K\right), (x,y,z) \in P \right\}$ according to the camera intrinsics $K$ and extrinsics $E$. 
Then the correspondence from LiDAR BEV to the camera image is defined as:
$\mathcal{M}_{p\rightarrow c}(i_p,j_p) \triangleq P_{c}$.

\textbf{Attention-based feature interaction.}
Once the cross-modality adjacency is dictated, we employ the attention mechanism to implement the inter-modal information exchange. Specifically, 
given an image feature as query $\boldsymbol{q}=\boldsymbol{h}_{c}^{[i_c,j_c]}$, 
its cross-modality neighbors $\mathcal{N}_q = \boldsymbol{h}_p^{\left[\mathcal{M}_{c\rightarrow p}(i_c,j_c)\right]}$,
are used as the \texttt{key} $\boldsymbol{k}$ and \texttt{value} $\boldsymbol{v}$ for cross-attention:
\begin{equation}
    f_{{\phi}_{p \rightarrow c}}
    \left( \boldsymbol{h}_c, \boldsymbol{h}_p \right)
    ^{[i_c,j_c]} = \sum_{\boldsymbol{k},\boldsymbol{v} \in \mathcal{N}_q}^{}\text{softmax}\left (  \frac{\boldsymbol{qk}}{\sqrt{d} }\right ) \boldsymbol{v},
\label{eq:local_attention}
\end{equation}
where $\boldsymbol{h}^{[i, j]}$ denotes indexing the element at location $(i,j)$ on the 2D representation $\boldsymbol{h}$, and $f_{{\phi}_{p \rightarrow c}}
\left( \boldsymbol{h}_c, \boldsymbol{h}_p \right)$ is {\em LiDAR-to-image representational interaction  (MMRI-I2L)}, yielding the image features augmented with the LiDAR information.

The other way around, given a LiDAR BEV feature point as a query $\boldsymbol{q}=\boldsymbol{h}_{p}^{[i_p,j_p]}$, we similarly obtain its cross-modality neighbors as $\mathcal{N}_q = \boldsymbol{h}_c^{\left[\mathcal{M}_{p\rightarrow c}(i_p,j_p)\right]}$.
The same process as Eq. (\ref{eq:local_attention}) can be applied for realizing {\em image-to-LiDAR representational interaction (MMRI-I2L)} 
$f_{{\phi}_{c \rightarrow p}} \left( \boldsymbol{h}_c, \boldsymbol{h}_p \right)$, which is illustrated in the Figure~\ref{fig:encoder} (b).

\textbf{LiDAR-guided cross-plane polar ray attention.} 
\label{sec:polarattn} 
To facilitate representational interaction between sparse LiDAR and dense image modalities, we need effective cross-modal representational enhancement. 
However, the aforementioned projection and sampling-based interaction operation employed in DeepInteraction~\cite{yang2022deepinteraction} suffers from sparse interaction with missing semantics due to the sparse nature of LiDAR data.
Although the consequential loss can be mitigated by incorporating complete image representation in the decoding process, it may still lead to insufficient supervision for the cross-plane matching process, resulting in suboptimal representation learning for the image-enhanced LiDAR BEV features. Additionally, this interaction's heavy reliance on precise LiDAR calibration could compromise the system's overall robustness.

Incorporating dense global context is conducive to further performance gains, particularly for image-to-BEV interaction as mentioned in~\cite{liu2022bevfusion}. 
Therefore, we introduce a new interaction mechanism, \ie, \textit{LiDAR-guided cross-plane attention} between the image column and BEV polar ray, inspired by~\cite{jiang2022polar}. This is designed to effectively leverage dense image features in representational interaction.
This module is inserted between the self-attention and the cross-attention of the Transformer layer described in Eq.~(\ref{eq:tranformer}). It enables our image-to-LiDAR representational interaction to effectively use the dense global context in image information while maintaining sparse local focus at the object level.

The new cross-attention operation leverages the inherent correspondence between the BEV polar ray and the camera image column. 
Instead of relying solely on learning-based cross-plane feature alignment as \cite{jiang2022polar}, our approach integrates LiDAR information as guidance. Specifically, for each camera $c$, we first transform $\boldsymbol{h}_{p}^{p\rightarrow p}$ into the polar coordinate system with origin $c$ and obtain $\boldsymbol{h}_{polar}\in \mathbb{R}^{R \times W \times C}$, where $W$ is the width of the image feature $\boldsymbol{h}_{c}$, and $R$ is the dimension of the radius. After transformation, the $i$-th polar ray in LiDAR BEV feature map, $\boldsymbol{h}_{polar}^{[:,i]}$, naturally corresponds to the $i$-th column in the image feature map $\boldsymbol{h}_{c}^{[:,i]}$. Once the camera parameters are fixed, the one-to-one correspondence between elements of the two sequences will become more stable and easier to learn. We leverage multi-head attention with sinusoidal position encoding to capture this pattern,
\begin{equation}
\begin{aligned}
\label{eq:crossplane}
    (\boldsymbol{h}_{polar}^{c\rightarrow polar})^{[:,i]} =
    MHA( 
    &Q=\boldsymbol{h}_{polar}^{[:,i]},\\
    &K=\boldsymbol{h}_{c}^{[:,i]},\\
    &V=\boldsymbol{h}_{c}^{[:,i]}
    ).
\end{aligned}
\end{equation}
$\boldsymbol{h}_{polar}^{c\rightarrow polar}$ is the LiDAR feature map enhanced by the image representation $\boldsymbol{h}_{c}$ and will be transformed back into the cartesian coordinate system for subsequent interaction. With the assistance of LiDAR information, this transformation is more tractable compared to those image-only approaches, which need to repeat the multi-head attention several times to spread image semantics to the correct depth.

Furthermore, we employ the flash attention~\cite{dao2022flashattention,dao2023flashattention2} to minimize the additional computation and memory overhead introduced by this module. 
The experimental results in Section~\ref{sec:experiments} demonstrate that this operation provides a beneficial dense context, which complements the original object-centric sparse interaction, thus significantly enhancing detection performance and enabling the extension to end-to-end planning.

\noindent \textit{A.3 Intra-modal representational learning (IML)}

Beyond directly incorporating information from heterogeneous modalities, intra-modal reasoning is helpful for more comprehensive integration of these representations. 
Therefore, in each layer of the encoder, we conduct intra-modal representational learning
complementary to multi-modal interaction. In this work, we utilize deformable attention~\cite{zhu2020deformable} for intra-modal representational learning, replacing the stand-alone attention~\cite{ramachandran2019stand} in the original version. Considering the scale variance introduced by perspective projection, interaction operation with a more flexible receptive field would be more reasonable than conducting cross-attention within fixed local neighbors as~\cite{yang2022deepinteraction}. This modification maintains the original efficient local computation while achieving a more flexible receptive field and facilitating the multi-scale interaction.

\noindent \textit{A.4 Efficient interaction with grouped sparse attention}
\label{sec:groupattn}

Given the inherent sparsity of point clouds, the number of LiDAR points varies within pillars depending on their position, and points within a single pillar are visible to no more than two cameras.
Therefore, to fully leverage the parallel computing capabilities of modern GPU devices during the image-to-LiDAR representational interaction, we first need to pad image tokens attended by each pillar to meet a fixed number and mask the invalid tokens within the attention. However, this brute-force approach will inevitably lead to substantial unnecessary computation and memory consumption.

To tackle this issue, we carefully examine the distribution of the number of valid image tokens per pillar and divide these pillars into several intervals $\mathcal{I} = \left\{ (N_{i}, N_{i+1}) \right\}_{i=0}^{N_{\mathrm{interval}}}$. Then we batchify pillars within each interval by padding the number of keys and values to the interval’s upper limit $ N_{i+1} $ for attention computations.
With careful selection of interval boundaries, this modification significantly reduces memory consumption with negligible impact on parallelism. Besides, it is computationally equivalent to the original implementation, as the padded tokens are masked during the attention process.

\subsection{Decoder: Multi-modal predictive interaction}
\begin{figure*}
    \centering
    \includegraphics[width=1.0\linewidth]{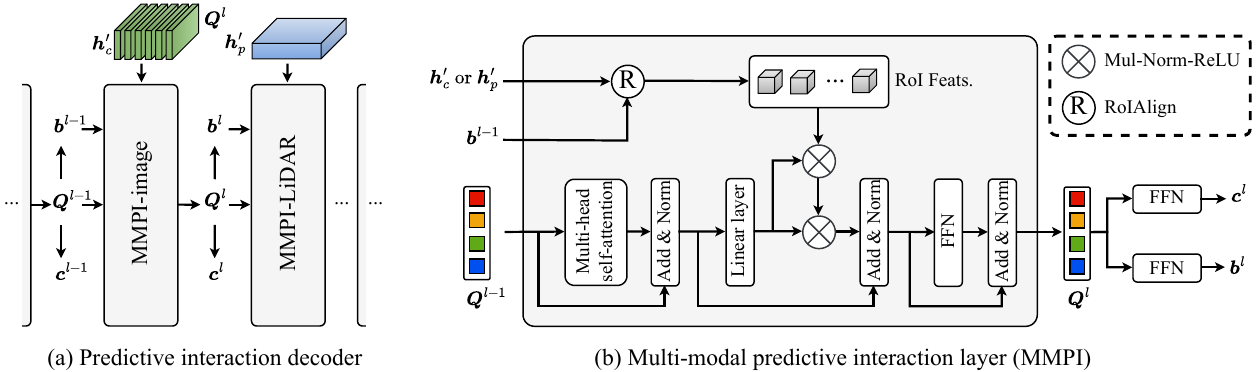}
    \caption{Illustration of our multi-modal predictive interaction.
    Our predictive interaction decoder \textbf{(a)} generates predictions via \textbf{(b)} progressively interacting with two modality-specific representations. 
    }
    \label{fig:decoder_layer}
    \vspace{-5mm}
\end{figure*}

\label{sec:decoder}
Beyond considering the multi-modal interaction at the representation level, we further introduce a decoder with {\em multi-modal predictive interaction}(MMPI) to unleash the modality-specifical information storage in separate representations and maximize their complementary effects in prediction.

As depicted in Figure~\ref{fig:decoder_layer}(a), our core idea is to enhance the 3D object detection of one modality conditioned on the other modality.
In particular, the decoder is built by stacking multiple {\em multi-modal predictive interaction layers}, within which {\em predictive interactions} are deployed 
to progressively refine the predictions by alternatively aggregating information from the enhanced image representation $\boldsymbol{h}_c^\prime$ and the enhanced BEV representation $\boldsymbol{h}_p^\prime$.
Similar to the decoder of DETR \cite{carion2020detr}, we cast the 3D object detection as a set prediction problem. 
Here, we define $N$ object queries $\left \{\boldsymbol{Q}_{n}\right \}^N_{n=1}$ which will transform into $N$ object predictions $\left \{ (\boldsymbol{b}_n, \boldsymbol{c}_n) \right \}^N_{n=1}$ through the decoder, where $\boldsymbol{b}_n$ and $\boldsymbol{c}_n$ denote the predicted bounding box and category decoded from the $n$-th query. To enable effective multi-modal interaction for model predictions, we propose \textit{multi-modal predictive interaction layer} to build the decoder.
For the $l$-th decoder layer, the set prediction is computed by taking the query embeddings $\left \{\boldsymbol{Q}_{n}^{(l-1)}\right \}^N_{n=1}$ and the predicted bounding boxes $\left \{\boldsymbol{b}_{n}^{(l-1)}\right \}^N_{n=1}$ from previous layer as inputs and 
enabling interaction with the enhanced image $\boldsymbol{h}_p^{\prime}$ or LiDAR $\boldsymbol{h}_c^\prime$ representations ($\boldsymbol{h}_c^{\prime}$ if $l$ is odd, $\boldsymbol{h}_p^{\prime}$ if $l$ is even).
We formulate the multi-modal predictive interaction layer (Figure~\ref{fig:decoder_layer}(b)) for specific modality as follows.

\textbf{MMPI on image representation.}
Taking as input 3D object proposals $\left \{\boldsymbol{b}_{n}^{(l-1)}\right \}^N_{n=1}$ and corresponding query embeddings $\left \{\boldsymbol{Q}_{n}^{(l-1)}\right \}^N_{n=1}$ produced by the previous layer, 
the current layer will leverage the image representation $\boldsymbol{h}_c^{\prime}$
for further prediction refinement. 
To integrate the previous predictions $\left \{\boldsymbol{b}_{n}^{(l-1)}\right \}^N_{n=1}$, 
we first extract $N$ Region of Interest (RoI)~\cite{he2017maskrcnn} features $\left \{ \boldsymbol{R}_{n} \right \}^N_{n=1}$ from the image representation $\boldsymbol{h}_c^{\prime}$, where $\boldsymbol{R}_{n} \in \mathbb{R}^{S \times S \times C}$ is the extracted RoI feature for the $n$-th query, $(S \times S)$ is the size of RoI, and $C$ is the number of channels. 
Specifically, for each 3D bounding box, we project it onto image representation $\boldsymbol{h}_c^{\prime}$ to get the 2D convex polygon and take the minimum axis-aligned circumscribed rectangle as its RoI.
We then design a multi-modal predictive
interaction operator
that first maps $\left \{\boldsymbol{Q}_{n}^{(l-1)}\right \}^N_{n=1}$ into the parameters of a series of $1 \times 1$ convolutions and then applies them consecutively on the RoI feature $\left \{ \boldsymbol{R}_{n} \right \}^N_{n=1}$;
Finally, the resulting feature will be used
to update object query $\left \{\boldsymbol{Q}_{n}^{l}\right \}^N_{n=1}$.

\textbf{MMPI on LiDAR representation.} This layer shares the same design as the above except that it takes as input LiDAR representation instead. With regards to the RoI for LiDAR representation, we project the 3D bounding boxes from the previous layer to the LiDAR BEV representation $\boldsymbol{h}_p^{\prime}$ and take the minimum axis-aligned rectangle. It is worth mentioning that due to the scale of objects in autonomous driving scenarios being usually tiny in the BEV coordinate frame, we enlarge the scale of the 3D bounding box by $2\times$ for RoI Align. The shape of RoI features cropped from the LiDAR BEV representation $\boldsymbol{h}_p^{\prime}$ is also set to be $S \times S \times C$. Here $C$ is the number of channels of RoI features and BEV representation. The multi-modal predictive interaction layer for LiDAR representation is stacked on its image counterpart.

For the prediction decoding, a feed-forward network is appended on the $\left \{\boldsymbol{Q}_{n}^{l}\right \}^N_{n=1}$ for each multi-modal predictive interaction layer to infer the classification score, locations, dimensions, orientations, and velocities. 
During training, the matching cost and loss function with the same form as in~\cite{bai2022transfusion} are applied to each layer.

\begin{table*}[ht]
\caption{Comparison with state-of-the-art methods for 3D object detection on the nuScenes \texttt{test} set.
\texttt{Metrics}: mAP(\%), NDS(\%).
\dag~denotes test-time augmentation is used. 
}
\scriptsize
\renewcommand\tabcolsep{3.0pt}
\renewcommand\arraystretch{1.2}
\renewcommand{\footnote}{\fnsymbol{footnote}} 
\small
\setlength{\abovecaptionskip}{0.0cm}
\setlength{\belowcaptionskip}{-0.45cm}
\centering

\begin{tabular}{l|c|cc|cc|cc}
\hline

\hline

\hline
\multirow{2}*{Method} & \multirow{2}*{Present at} &  \multicolumn{2}{c}{Backbones}\vline & \multicolumn{2}{c}{\textit{validation}}\vline & \multicolumn{2}{c}{\textit{test}} \\ 
& & Image & LiDAR & ~~mAP$\uparrow$~~ & ~~NDS$\uparrow$~~ & ~~mAP$\uparrow$~~ & ~~NDS$\uparrow$~~ \\
\hline

\hline

TransFusion~\cite{bai2022transfusion} & CVPR'22  & R50 & VoxelNet & 67.5 & 71.3 &  68.9 & 71.6 \\

MSMDFusion~\cite{jiao2023msmdfusion} & CVPR'23 & R50 & VoxelNet & 69.3 & 72.1 & 71.5 & 74.0 \\

SparseFusion~\cite{xie2023sparsefusion} & ICCV'2023  & R50 & VoxelNet & 70.5 & 72.8 & 72.0 & 73.8 \\

FUTR3D~\cite{chen2022futr3d} & arXiv'22 & R101 & VoxelNet  & 64.5 & 68.3 & - & - \\

PointAugmenting~\cite{wang2021pointaugmenting}\dag & CVPR'2021  & DLA34 & VoxelNet & - & - & 66.8 & 71.0  \\

MVP~\cite{Yin2021MVP} & NeurIPS'21  & DLA34 & VoxelNet & 67.1 & 70.8 & 66.4 & 70.5\\

AutoAlignV2~\cite{chen2022autoalign} & ECCV'22  & CSPNet & VoxelNet & 67.1 & 71.2  & 68.4 & 72.4 \\

BEVFusion~\cite{liang2022bevfusion} & NeurIPS'22  & Swin-Tiny & VoxelNet  & 67.9 & 71.0 & 69.2 & 71.8 \\

BEVFusion~\cite{liu2022bevfusion} & ICRA'23  & Swin-Tiny & VoxelNet & 68.5 & 71.4  & 70.2 & 72.9 \\

SparseFusion~\cite{li2024sparsefusion} & arXiv'24 & Swin-Tiny & VoxelNet & 68.7 & 70.6 & 70.1 & 72.7 \\

ContrastAlign~\cite{song2024contrastalign} & arXiv'24 & Swin-Tiny & VoxelNet & 70.3 & 72.5 & 71.8 & 73.8 \\

CMT~\cite{yan2023cross} & ICCV'23  & VOVNet & VoxelNet & 70.3 & 72.9  & 72.0 & 74.1 \\

UniTR~\cite{wang2023unitr} & ICCV'23  & DSVT~\cite{wang2023dsvt} & DSVT & 70.5 & 73.3  & 70.9 & \textbf{74.5} \\

FSF~\cite{li2024fully} & TPAMI'24  & HTC & FSD~\cite{fan2022fully} & 70.4 & 72.7 & 70.6 & 74.0 \\

\rowcolor[gray]{.9} 
DeepInteraction & NeurIPS'22  & R50 & VoxelNet  & 69.9 & 72.6 & 70.8 & 73.4 \\

\rowcolor[gray]{.9} 
DeepInteraction++ & Submission & Swin-Tiny & VoxelNet  & \textbf{70.6} & \textbf{73.3} & \textbf{72.0} & 74.4 \\
\hline

\hline

\hline
\end{tabular}
\label{tab:nuscene_test}
\vspace{-3mm}
\end{table*}

\section{DeepInteraction++ for end2end autonomous driving}
\label{e2d}

To further demonstrate the scalability and superiority, we extend our DeepInteraction++ to an end-to-end multi-task framework, simultaneously resolving scene perception, motion prediction, and ego-planning tasks. Instead of involving numerous sub-tasks of the driving scenario, we affiliate three additional downstream tasks (including map segmentation, prediction, and planning) following VAD~\cite{jiang2023vad}, a relatively lightweight framework. Hence, our end-to-end variant can effectively alleviate the memory overhead caused by the complicated interaction encoder and further unleash the multi-task capabilities benefiting from multi-modal representations.

We employ extra task heads besides the existing detection head to form the end-to-end framework, constituted by a segmentation head for map segmenting, a prediction head to estimate the motion status of detected objects, and a planning head to provide a final action plan for ego vehicles. Considering that the feature maps from BEV and the surrounding view are utilized for deep interactive decoding, we make some modifications to leverage this advantage. First, compared to LiDAR points, the image context is more discriminative for the map representation, and massive point information might reversely cause confusion. Hence, we project the surrounding-view features onto BEV by LSS~\cite{philion2020lift} and then propagate them into the map segmentation head. Subsequently, the prediction and planning heads take as input the results generated by detection and segmentation, processing them with standard Transformer decoders.

\begin{figure}[t]
    \centering
    \includegraphics[width=1\linewidth]{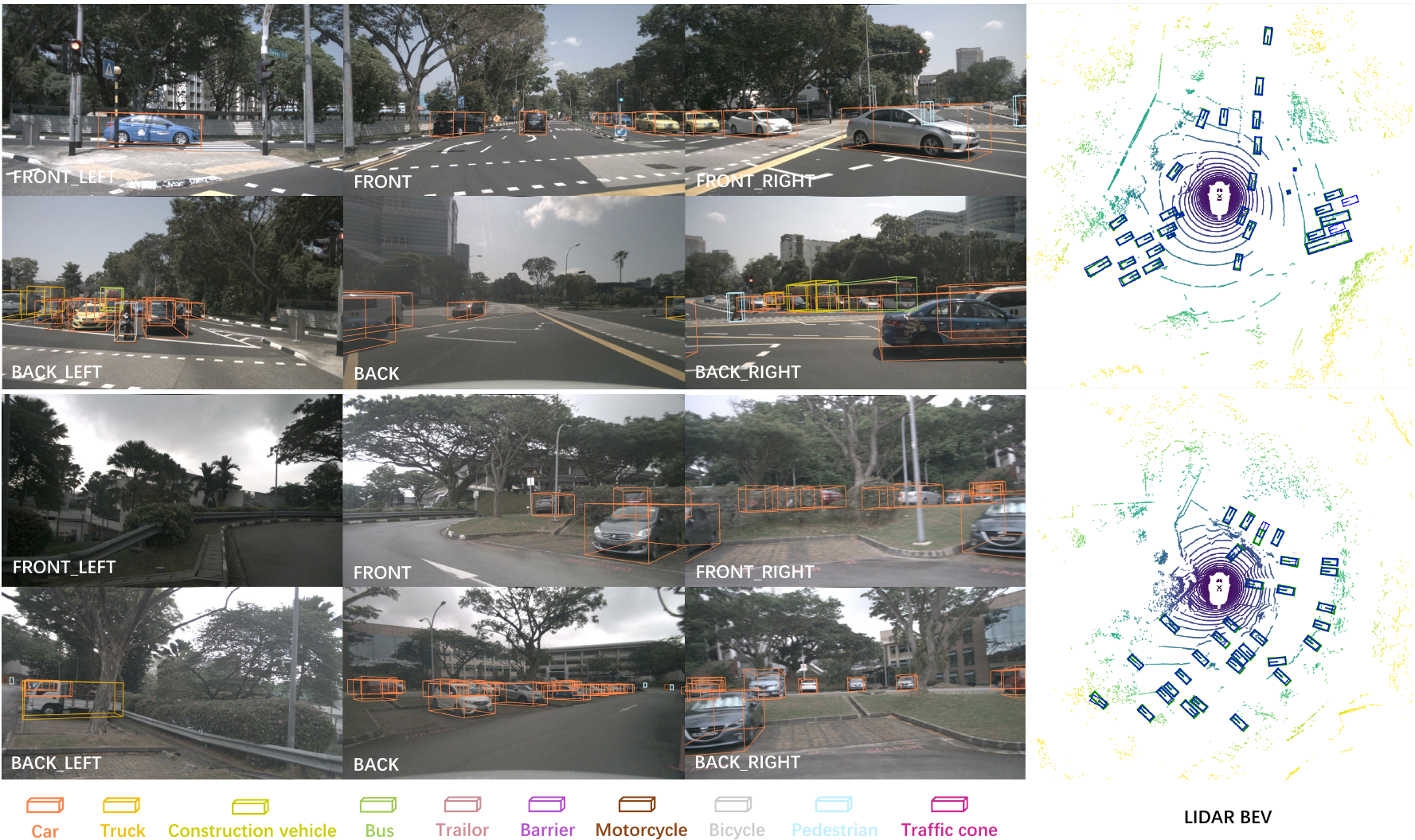}
    \caption{Qualitative results on nuScenes {\tt val} set. In LiDAR BEV (right), green boxes are the ground-truth and blue boxes are the predictions. Best viewed when zooming in.}
    \label{fig:quanlitative_result_nuscenes}
    \vspace{-5mm}
\end{figure}
\section{Experiments}
\label{sec:experiments}

\subsection{Experimental setup}

{\noindent \textbf{Dataset.}}
We evaluate our approach on the nuScenes dataset~\cite{caesar2020nuscenes}, which provides point clouds from 32-beam LiDAR and images with a resolution of $1600 \times 900$ from 6 surrounding cameras. It contains 1000 scenes and is officially split into \texttt{train/val/test} set with 700/150/150 scenes, where each sequence is roughly 20 seconds long and annotated every 0.5 seconds, 
For the 3D object detection task, 1.4M objects in various scenes are annotated with 3D bounding boxes and classified into 10 categories: car, truck, bus, trailer, construction vehicle, pedestrian, motorcycle, bicycle, barrier, and traffic cone.

{\noindent \textbf{Metric.}} For evaluation,
we leverage mean average precision (mAP)~\cite{everingham2010pascal} and nuScenes detection score (NDS)~\cite{caesar2020nuscenes} as score metrics to measure 3D detection performance. Specifically, we compute mAP by averaging over the distance thresholds of 0.5m, 1m, 2m, and 4m across 10 classes.
NDS is a weighted average of mAP and other attribute metrics, including translation, scale, orientation, velocity, and other box attributes. 

\begin{table}[t]
\caption{
Run time comparison measured on an NVIDIA RTX A6000 GPU. If not specified with $^{\star}$, the performance is evaluated on the nuScenes {\tt val} set. 
}
\renewcommand\tabcolsep{5.2pt}
\renewcommand\arraystretch{1.2}
\small
\begin{center}
 
\begin{tabular}{l|ccc}
\hline

\hline

\hline
{Method} & {mAP$\uparrow$} & {NDS$\uparrow$} & {FPS$\uparrow$} \\
\hline

\hline
PointAugmenting~\cite{wang2021pointaugmenting} & 66.8$^{\star}$ & 71.0$^{\star}$ & 2.8 \\
TransFusion~\cite{bai2022transfusion} & 67.5 & 71.3 & \textbf{5.5}  \\
FUTR3D~\cite{chen2022futr3d} & 64.2 & 68.0 & 2.3  \\
CMT~\cite{yan2023cross} & 70.3 & 72.9 & 3.3 \\
\rowcolor[gray]{.9}
DeepInteraction  & 69.9 & 72.6 & 3.1 \\
\rowcolor[gray]{.9} 
DeepInteraction++ & \textbf{70.6} & \textbf{73.3} & 3.9  \\
\hline

\hline

\hline
\end{tabular}
\end{center}
\label{tab:efficiency}
\vspace{-5mm}
\end{table}

\begin{table*}[t]
\caption{
Quantitative comparison of detection performance between DeepInteraction and DeepInteraction++ under different image backbones. The results are evaluated on the nuScenes \texttt{val} split. 
}
\small
\centering
\begin{tabular}{l|c|ccccccc}
\hline

\hline

\hline
\multirow{2}*{Method} & image & \multirow{2}*{NDS$\uparrow$} & \multirow{2}*{mAP$\uparrow$} & \multirow{2}*{mATE$\downarrow$} & \multirow{2}*{mASE$\downarrow$} & \multirow{2}*{mAOE$\downarrow$} & \multirow{2}*{mAVE$\downarrow$} & \multirow{2}*{mAAE$\downarrow$} \\

 & backbone & & & & & & & \\
\hline

\hline
DeepInteraction  & \multirow{2}*{R50}  & \textcolor{blue}{72.6} & \textcolor{blue}{69.9} & \textcolor{blue}{26.7} & \textbf{\textcolor{blue}{25.0}} & \textcolor{blue}{27.6} & \textcolor{blue}{24.8} & \textcolor{blue}{18.9} \\
DeepInteraction++ & & \textbf{\textcolor{blue}{72.9}} & \textbf{\textcolor{blue}{70.1}} & \textbf{\textcolor{blue}{26.5}} & \textcolor{blue}{25.1} & \textbf{\textcolor{blue}{26.6}} & \textbf{\textcolor{blue}{24.5}} & \textbf{\textcolor{blue}{18.9}} \\
\hline
DeepInteraction & \multirow{2}*{Swin-Tiny} & \textcolor{blue}{72.6} & \textcolor{blue}{70.0} & \textcolor{blue}{27.0} & \textbf{\textcolor{blue}{25.2}} & \textcolor{blue}{28.1} & \textcolor{blue}{24.5} & \textcolor{blue}{18.9} \\
DeepInteraction++ & & \textbf{\textcolor{blue}{73.3}} & \textbf{\textcolor{blue}{70.6}} & \textbf{\textcolor{blue}{26.8}} & \textcolor{blue}{25.3} & \textbf{\textcolor{blue}{26.2}} & \textbf{\textcolor{blue}{23.4}} & \textbf{\textcolor{blue}{18.6}} \\
\hline

\hline

\hline
\end{tabular}
\label{tab:journal_main}
\vspace{-5mm}
\end{table*}

\subsection{Implementation details}
\label{sec:details}

{\noindent \textbf{Model.}} We implement our model framework based on the public codebase \textit{mmdetection3d}~\cite{mmdet3d2020}. 
Following TransFusion~\cite{bai2022transfusion}, we initialize our image backbone from the instance segmentation model {\em Cascade Mask R-CNN}~\cite{cai2019cascade} pretrained on COCO~\cite{lin2014coco} and nuImage~\cite{caesar2020nuscenes}. 
For DeepInteraction and DeepInteraction++, we employ widely used ResNet-50~\cite{he2016deep} and Swin-Tiny~\cite{liu2021Swin} as the default backbone for image modality, respectively.
To save the computation cost, we downscale the input image size to half and freeze the parameters of the image backbone during training.
For a fair comparison with other alternates, we set the voxel size to $(0.075m, 0.075m, 0.2m)$, and the detection range to $[-54m, 54m]$ for $X$ and $Y$ axis and $[-5m, 3m]$ for $Z$ axis in the default configuration. 
For the multi-modal interactive modules, we build the encoder 
by stacking two representational interaction layers and the decoder with five cascaded predictive interaction layers.
We set the query number to 200 for training and employ the same query initialization strategy as Transfusion~\cite{bai2022transfusion}. During testing, we adapt the number of queries to 300 and 400 for DeepInteraction and DeepInteraction++, respectively, to achieve the best performance. 
Note that, test-time augmentation and model ensemble tricks are not explored in this work.

{\noindent \textbf{Training.}} Following the common practice, we adopt several random data augmentations, including rotation with a range of $r \in \left[ -\pi/4,\pi/4 \right]$, scaling with a factor of $r \in \left[ 0.9,1.1 \right]$, translation with standard deviation 0.5 in three axes, and horizontal flipping.  We use the class-balanced re-sampling in CBGS~\cite{zhu2019class} to balance the class distribution for the nuScenes dataset. Following~\cite{bai2022transfusion}, we adopt a two-stage training recipe. 
We take TransFusion-L~\cite{bai2022transfusion} as our \texttt{LiDAR-only baseline} and train LiDAR-image fusion modules for 6 and 9 epochs with a batch size of 16 on 8 NVIDIA A6000 GPUs for DeepInteraction and DeepInteraction++, respectively.
During training, we use the Adam optimizer with a one-cycle learning rate policy, with a max learning rate of $1 \times 10^{-3}$, weight decay 0.01, and momentum 0.85 to 0.95 as in CBGS~\cite{zhu2019class}.

\subsection{Comparison to the state of the arts}
\label{sec:sota}

{\noindent \textbf{Main results.}} We compare with state-of-the-art alternatives on both the \texttt{val} and \texttt{test} splits of nuScenes dataset. 
As shown in Table~\ref{tab:nuscene_test}, 
our vanilla DeepInteraction has surpassed all its prior arts under the same settings by a considerable margin, and our DeepInteraction++ achieves new state-of-the-art performance with the improved architectural design.
Notably, compared to Transfusion~\cite{bai2022transfusion}, which is a representative uniliteral fusion baseline, our DeepInteraction provides a significant performance gain of 2.4\% mAP and 1.3\% NDS using the same modality-specific backbone and training recipe,
verifying the advantages of our multi-modal interaction approach. 
We provide the per-category results in Table~\ref{tab:nuscenes_val}. The qualitative results are shown in Figure~\ref{fig:quanlitative_result_nuscenes}.

Our DeepInteraction++ by default employs a stronger image backbone.
To demonstrate that the improvements brought by the revised architecture are consistent and essential, we additionally provide a systematic and comprehensive comparison between DeepInteraction and DeepInteraction++ under the same image backbone on the nuScenes {\tt val} set.
The results in Table~\ref{tab:journal_main} suggest that DeepInteraction++ with a more meticulously designed architecture consistently beats the baseline across most metrics under all settings while adhering to the same hierarchical modality interaction build.

We ascribe the performance gain to two aspects: (1) The standard Transformer architecture with enhanced intra-modal learning provides a smoother gradient backpropagation path and a more flexible receptive field than the naive design in the conference version, enabling more effective optimization.
(2) The LiDAR-guided cross-plane polar ray attention effectively utilizes the dense context in the image feature, providing a beneficial supplement to the object-centric sparse interaction in the Image-to-LiDAR representational interaction. In the following sections, rigorous ablation experiments will further substantiate these claims.

{\noindent \textbf{Runtime.}} We compare the inference speed of all methods on NVIDIA RTX A6000 GPU.
As shown in Table~\ref{tab:efficiency}, our method achieves the best performance with faster inference speed than alternative painting-based~\cite{wang2021pointaugmenting} and query-based~\cite{chen2022futr3d,yan2023cross} fusion approaches.
This demonstrates that our method achieves a better trade-off between performance and efficiency. Specifically, feature extraction for multi-view high-resolution camera images contributes the most of the overall latency in a multi-modal 3D detector as verified in~\cite{wang2021pointaugmenting}. Our interaction modules are built with a relatively lighter model architecture that offers better running speed.
From Figure~\ref{fig:ablation_for_head_layer}, we observe that increasing the number of decoder layers only brings negligible extra latency, which concurs with the same conclusion.

\subsection{Ablation studies}

In this section, we first conduct ablations on DeepInteraction++ to study the effectiveness of our core model, {\em modality interaction}, and important design choices. Subsequently, we will provide a clear improvement trajectory from DeepInteraction to DeepInteraction++.

\subsubsection{Ablations of the modality interaction}
\begin{table}[ht]
\caption{Effects of modality interaction. We ablate each modality at different stages of interaction. ``I2L'' and ``L2I'' denote the image-to-LiDAR and LiDAR-to-Image representational interaction, respectively. ``L'' and ``I'' indicate the used modality in the decoder. All experiments are conducted on the DeepInteraction framework.}
\centering
\begin{tabular}{c|cc|cc|ccc}
\hline

\hline

\hline
& \multicolumn{2}{c}{Encoder}\vline & \multicolumn{2}{c}{Decoder}\vline & ~\multirow{2}*{mAP$\uparrow$}~ & ~\multirow{2}*{NDS$\uparrow$}~ & ~\multirow{2}*{FPS$\uparrow$}~ \\
& I2L & L2I & ~L~ & ~I~ &  &  &  \\
\hline

\hline
a) & \checkmark & & \checkmark &  & 68.9 & 71.9 & 5.6 \\
b) & \checkmark & & \checkmark & \checkmark & 69.4 & 72.5 & 4.8 \\
c) & \checkmark & \checkmark &  \checkmark & & 69.2 & 72.2 & 3.3 \\
d) & \checkmark & \checkmark & \checkmark & \checkmark & \textbf{69.9} & \textbf{72.6} & 3.1 \\
\hline

\hline

\hline
\end{tabular}
\label{tab:decoder_space}
\end{table} 
\quad

{\noindent \textbf{Effects of the representational interaction. }}
To demonstrate the superiority of our multi-modal representational interaction, we compare it with a degraded baseline, which does not iteratively refine the image features during the representational interactions.
For a fair comparison, both methods use the same number of encoder layers as well as the same decoder.
As shown in Table~\ref{tab:decoder_space} a) and c), our representational interaction is more effective than the unilateral fusion alternatives. Besides, we compare the representative Transfusion~\cite{bai2022transfusion} with the conventional modality fusion strategy in Tables~\ref{tab:backbone_pillar}, indicating the advantages of our bilateral modality interaction strategy.

{\noindent \textbf{Effects of the predictive interaction. }}
In Table~\ref{tab:decoder_space}, we evaluate the performance of using different representations/modalities in model decoding. Variants c) and d) compare the complete MMPI using both representations alternatively and using LiDAR-only representation in all decoder layers.
The results demonstrate the advantage of interacting with both modalities in the decoding stage. This suggests that even after sufficient mutual enhancement through a well-designed representational interaction mechanism, image representations still contain information with unique benefits for prediction.

{\noindent \textbf{Impact of the form of image representation. }}
\begin{table}[t]
\caption{
{\color{blue}
Ablation on the image representation. All experiments are based on our DeepInteraction++ framework.}
}
\label{tab:ablation_form}
\centering
\begin{tabular}{c|c|cc}
\hline

\hline

\hline
& Image feature form & ~~mAP$\uparrow$~~ & ~~NDS$\uparrow$~~ \\
\hline

\hline
a) & Fused & 69.2 & 72.1 \\
b) & BEV & 69.4 & 72.2 \\
c) & Perspective & \textbf{70.3} & \textbf{73.0} \\
\hline

\hline

\hline
\end{tabular}
\vspace{-5mm}
\end{table}
{\color{blue}
During the interactions in both the encoder and decoder, we consistently utilize perspective-form image features instead of converting them to the 3D space beforehand. This design choice is based on two key considerations:
(i) With the assistance of LiDAR point clouds, perspective image representation can already achieve precise interaction with LiDAR BEV features, limiting the potential benefits of lifting them into 3D space before the fusion encoder. 
As indicated in~\cite{yan2023cross,liu2022petr}, maintaining perspective image features is sufficient for 3D object detection task.
(ii) Due to the sparsity of LiDAR data and potential misalignment, this transformation process may be inaccurate and bring irreversible information loss.

To validate this, the comparison of different image representations is demonstrated in Table~\ref{tab:ablation_form}.
The BEV-form image representation is scattered by the perspective image feature using $\mathcal{M}_{c\rightarrow p}$. 
In a), the BEV image representation is concatenated with the LiDAR representation and fed into a single-stream Transformer encoder with only IML. In b), the BEV image representation is still kept separate from the LiDAR representation, while the cross-modal interaction in the encoder is replaced by simple deformable attention between two spatially aligned representations. The polar attention is disabled in both b) and c) for a fair comparison.
It can be observed that although b) is slightly better than directly fusing them, it still lags significantly behind c) where the image features are kept in perspective. These observations corroborate the rationale behind our design choice.
}

\subsubsection{Ablations on the encoder}
\begin{table}[t]
\caption{
\color{blue}
Ablation on the encoder design. \texttt{IML}: Intra-modal learning; \texttt{MMRI}: Multi-modal representational interaction. All experiments are based on our DeepInteraction++ framework.
}
\label{tab:abliation_interaction_on_neck}
\centering
\begin{tabular}{c|ccc|cc}
\hline

\hline

\hline
\# of encoder layers & IML & MMRI & Polar Attn. & mAP$\uparrow$ & NDS$\uparrow$ \\
\hline

\hline
w/o & & & & 67.7 & 71.7 \\
 \hline
1 & \checkmark & \checkmark & \checkmark & 70.0 & 72.9 \\
 \hline
 & \checkmark & & & 68.2 & 71.9 \\
 &  & \checkmark & & 70.0 & 72.8 \\
 &  &  & \checkmark & 69.7 & 72.5 \\
2 &  & \checkmark & \checkmark & 70.4 & 73.0  \\
 & \checkmark & \checkmark & & 70.3 & 73.0 \\
 & \checkmark &  & \checkmark & 69.9 & 72.6 \\
 & \checkmark & \checkmark & \checkmark & \textbf{70.6} & \textbf{73.3} \\
\hline

\hline

\hline
\end{tabular}
\end{table}
\begin{table}[t]
\caption{
{\color{blue}
Ablation on the Polar Attention on nuScenes and Waymo dataset
based on DeepInteraction++.}
}
\label{tab:polarwaymo}
\centering
\begin{tabular}{c|cc|cc}
\hline

\hline

\hline
& \multicolumn{2}{c|}{nuScenes}&  \multicolumn{2}{c}{Waymo (1/5 train)} \\

& ~~mAP$\uparrow$~~ & ~~NDS$\uparrow$~~ & ~~L2 mAP$\uparrow$~~ & ~~L2 mADH$\uparrow$~~ \\
\hline

\hline
w/o polar & 70.32 & 73.02 & 66.95 & 61.68 \\
w/ polar & \textbf{70.63} & \textbf{73.27} & \textbf{67.02} & \textbf{61.70} \\
\hline

\hline

\hline
\end{tabular}
\vspace{-5mm}
\end{table}
\quad

{\noindent \textbf{Design choices in the representational interaction encoder.}}
{
\color{blue}
The first row of Table~\ref{tab:abliation_interaction_on_neck} presents the result of the model without encoder, \ie, two modality-specific representations extracted independently from different backbones are directly fed into the decoder. 
Although this setting has already surpassed LiDAR-only baseline by a considerable margin, there is still a huge performance gap between it and configurations in other rows, underscoring the necessity of representational fusion between heterogeneous modalities for high-performance 3D detection.
To investigate exactly where these improvements come from, we ablate the multi-modal representational interaction (MMRI), intra-modal representational learning (IML), and LiDAR-guided cross-plane polar attention (Polar Attn.) in the encoder with various numbers of layers. 

We can draw several observations from Table~\ref{tab:abliation_interaction_on_neck}: 
(i) All three components contribute to the performance, while the MMRI and Polar Attn. play more critical roles since they introduce essential inter-modal information exchange. 
(ii) Stacking more encoder layers is essentially better than the shallow interaction.
(iii) While the polar ray attention brings considerable improvement to MMRI, it is insufficient to replace the role of MMRI when used independently. A plausible reason is that although it offers beneficial global context to the original object-centric sparse interaction, it is difficult to provide the precise interaction on its own, which is crucial for 3D object detection. We also evaluate this component on Waymo Open Dataset~\cite{sun2020scalability}, as shown in Table~\ref{tab:polarwaymo}. It demonstrates that indeed, polar attention is more effective for sparse LiDAR datasets than more dense cases (e.g., Waymo Open Dataset).
}

\begin{figure}[t]
    \centering 
    \setlength{\tabcolsep}{1.0pt}
    \begin{tabular}{ccc}
    \raisebox{4\height}{\scriptsize(a)} & \includegraphics[height=.2\linewidth,clip]{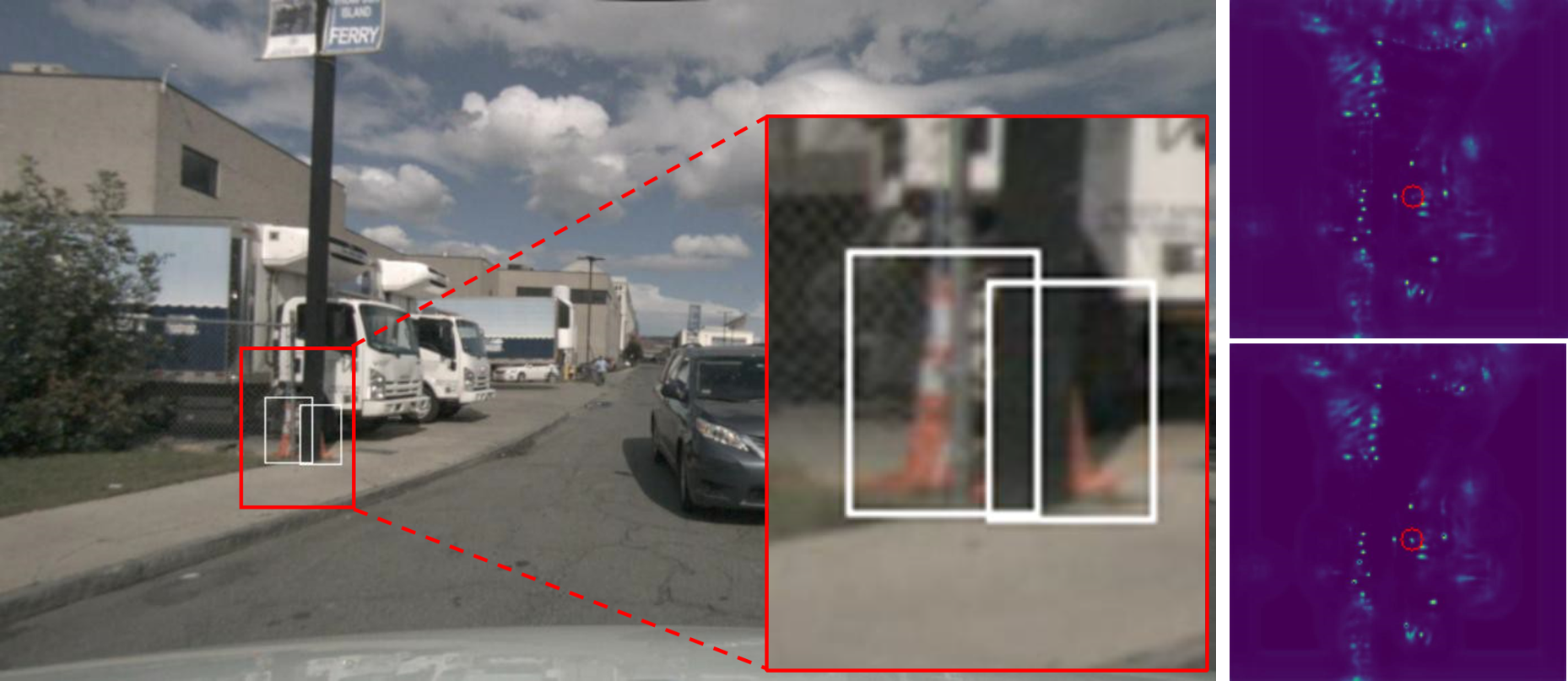}
    &
    \includegraphics[height=.2\linewidth,clip]{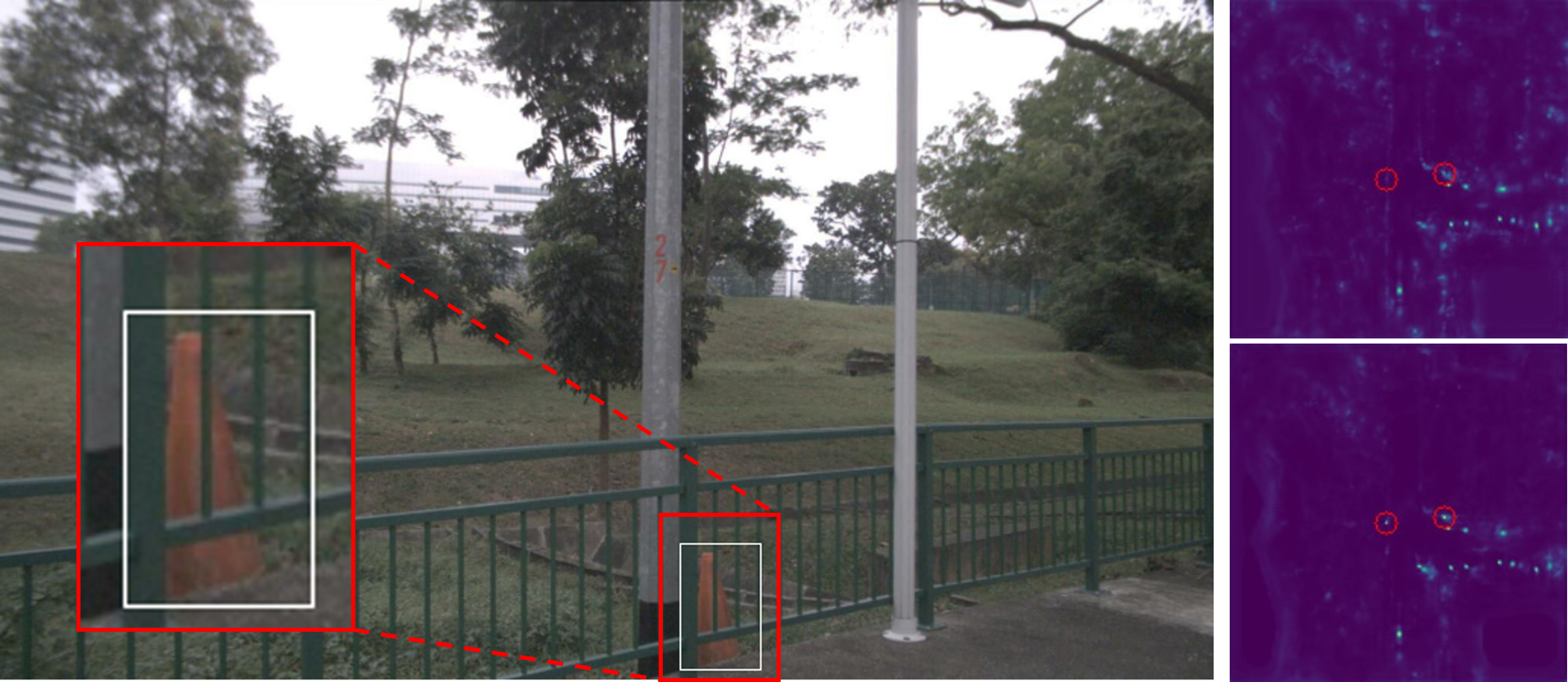}
\\
    \raisebox{4\height}{\scriptsize(b)} & \includegraphics[height=.2\linewidth,clip]{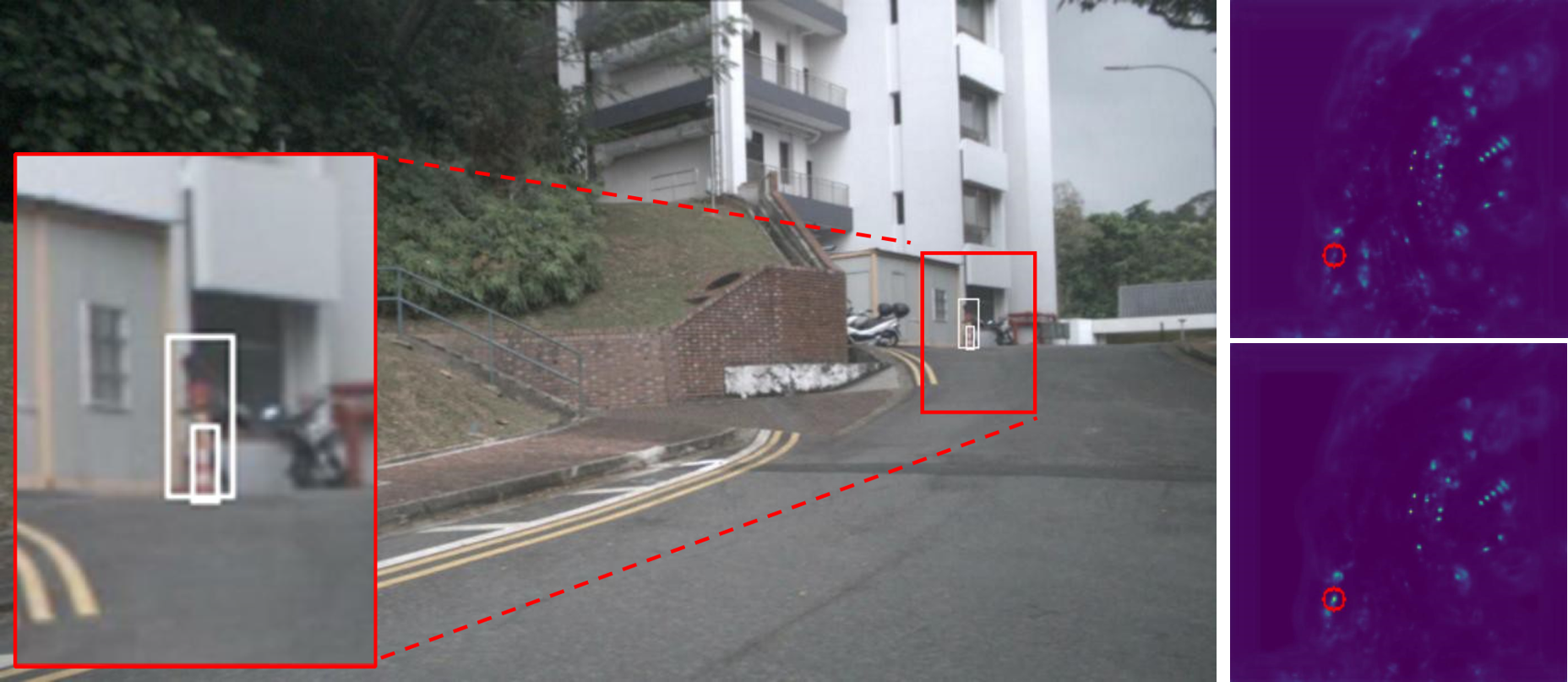}
    &
    \includegraphics[height=.2\linewidth,clip]{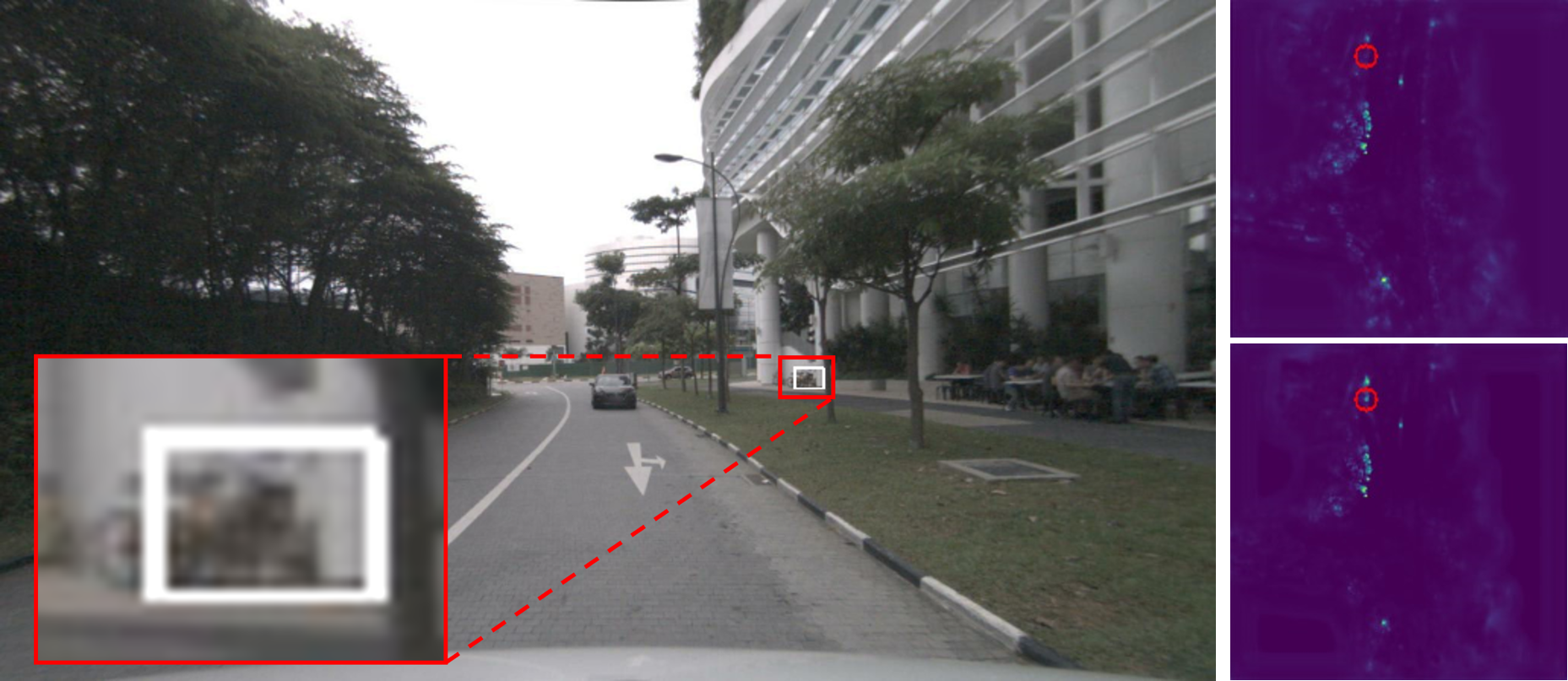}
\\
    \raisebox{5\height}{\scriptsize(c)} & \includegraphics[height=.2\linewidth,clip]{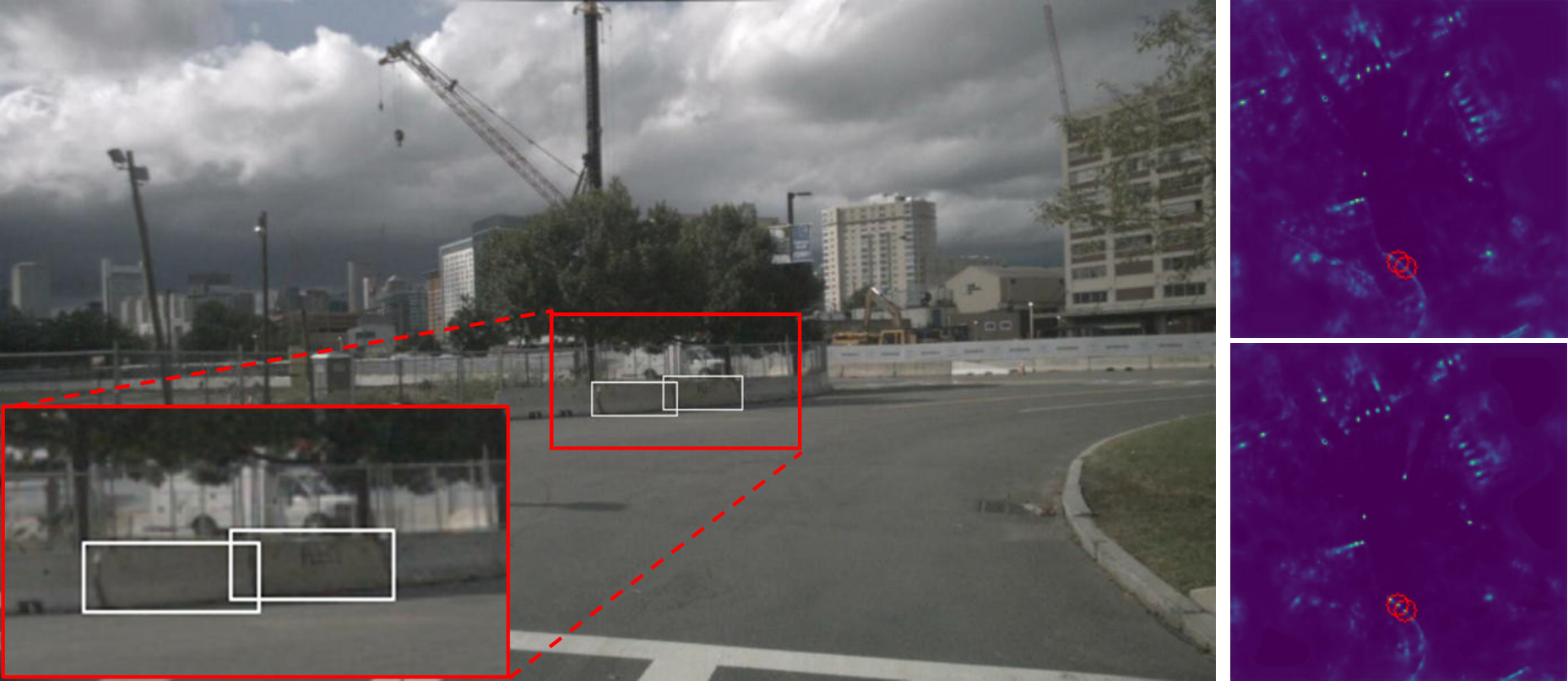}
    &
    \includegraphics[height=.2\linewidth,clip]{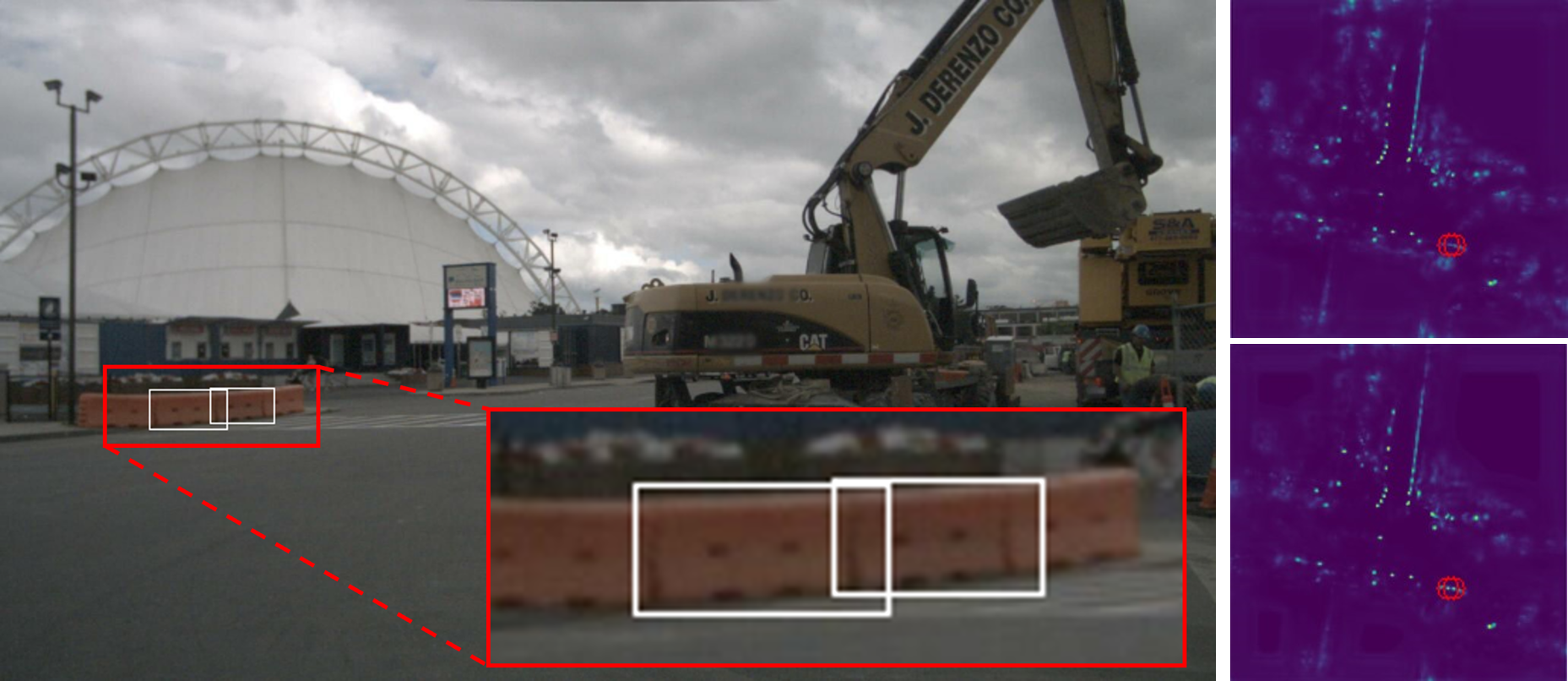}
\\

    \end{tabular}
\caption{
    Illustrations of the heatmaps predicted from BEV representations \textit{before (top)} and \textit{after (bottom)} representational interactions. All samples are from the nuScenes {\tt val} split. 
    \textbf{(a)} Occluded tiny objects. 
    \textbf{(b)} Small objects at long distance. 
    \textbf{(c)} Adjacent barriers connecting together in LiDAR point clouds thus difficult to discriminate 
    without the help of visual clues.
    }
    \label{fig:heatmap_vis_1}
\end{figure}
{\noindent \textbf{Qualitative results of representational interaction.}}
To gain more insight into the effect of our representational interaction, we visualize the predicted heatmaps of several challenging cases in the nuScenes dataset. 
From Figure~\ref{fig:heatmap_vis_1}, we can find that 
some objects will be neglected without the assistance of our representational interaction.
The locations of these objects are highlighted by red circles in the heatmap and white bounding boxes in the RGB image below.
Concretely, sample (a) suggests that camera information is helpful in recovering partially occluded tiny objects with few LiDAR points.
The sample (b) shows a representative case where some distant objects can be successfully recognized with the help of visual information.
From sample (c), we can observe that the centers of some barriers yield a more distinct activation in the heatmap after representational interaction. This is probably due to that it is too difficult to locate the boundaries of several consecutive barriers from LiDAR point clouds only.

\subsubsection{Ablations on the decoder}
\begin{table}[ht]
\caption{Ablation on the decoder design. We compare the performance between different types of operation employed for the interaction in decoding.}
\centering
\begin{tabular}{l|cc|cc}
\hline

\hline

\hline
Model & LiDAR & Image & ~mAP$\uparrow$~ & ~NDS$\uparrow$~ \\
\hline

\hline
& DETR~\cite{carion2020detr} & DETR & 68.6 & 71.6  \\
DeepInteraction & DETR & MMPI & 69.3 & 72.1  \\
& MMPI & MMPI & \textbf{69.9} &  \textbf{72.6}  \\
\hline
& DETR & DETR & 69.7 & 72.4  \\
DeepInteraction++ & DETR & MMPI & 70.2 & 72.7  \\
& MMPI & MMPI & \textbf{70.6} & \textbf{73.3}  \\
\hline

\hline

\hline
\end{tabular}
\label{tab:ablation_for_interaction_of_head}
\vspace{-5mm}
\end{table}
\begin{table}[ht]
\caption{Ablation on the number of queries used for 3D detection. All experiments are conducted on the DeepInteraction framework.}
\centering
\begin{tabular}{c|c|cc}
\hline

\hline

\hline
~~~~~Train~~~~ & ~Inference~ &  ~mAP~ & ~NDS~ \\ 
\hline

\hline
& 200 & 69.9 & 72.6  \\
200 & 300 & \textbf{70.1} &  \textbf{72.7}  \\ 
& 400 & 70.0 & 72.6 \\
\hline
& 200 & 69.7 & 72.5 \\
300 & 300 & 69.9 &  72.6  \\ 
& 400 & 70.0 & 72.6 \\
\hline

\hline

\hline
\end{tabular}
\label{tab:num_of_query}

\end{table}
\begin{figure}[ht]
    \centering
    \includegraphics[width=0.8\linewidth]{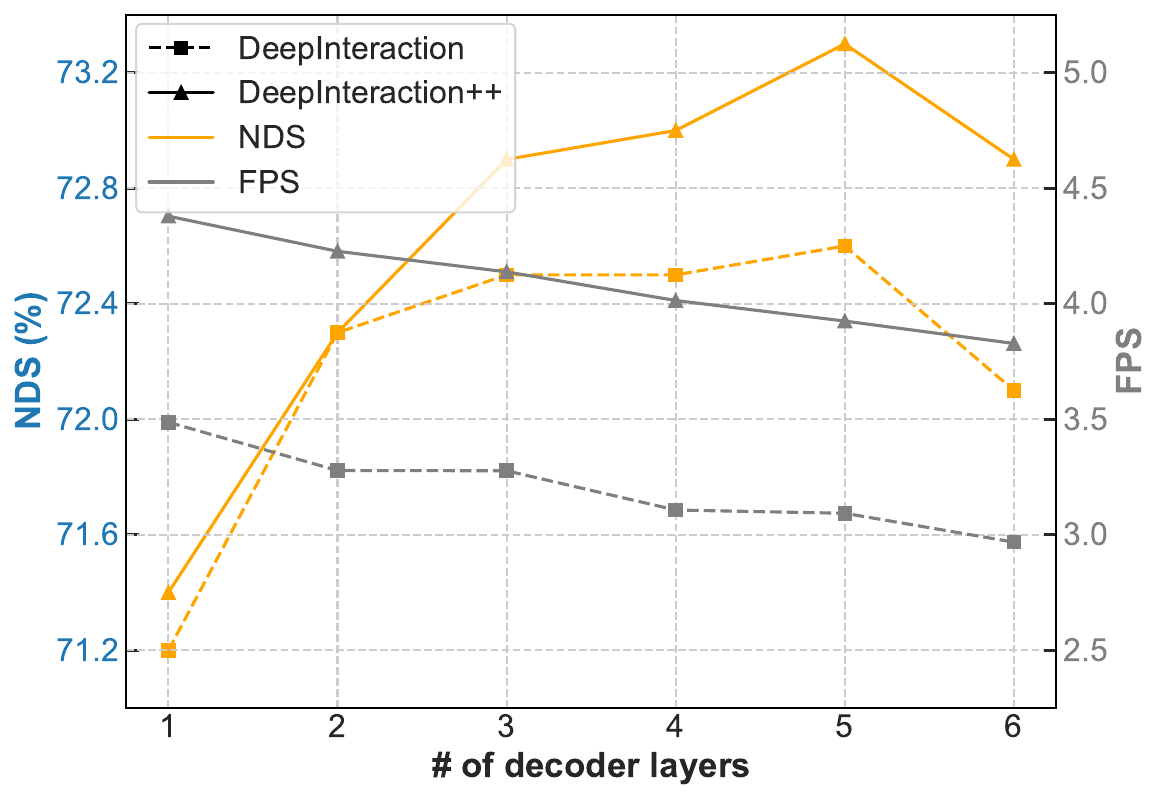}
    \caption{
    3D detection performance with the different number of decoder layers.  
    }
    \label{fig:ablation_for_head_layer}
    \vspace{-5mm}
\end{figure} 

\quad

{\noindent \textbf{Multi-modal predictive interaction layer vs. standard DETR~\cite{carion2020detr} prediction.}}
In Table~\ref{tab:ablation_for_interaction_of_head}, we evaluate the effect of the design for predictive interaction by comparing our multi-modal predictive interaction (MMPI) with standard
DETR~\cite{carion2020detr} decoder layer.
Note the latter setting means the vanilla cross-attention is used to aggregate multi-modal information as in Transfusion~\cite{bai2022transfusion}.
We further test a mixing design:
using the cross-attention for aggregating features in LiDAR representation and MMPI for image representation.
{
\color{blue} 
Note that this ablation only verifies the module advantage of MMPI layers over DETR layers, where the interaction order with each modality is consistent in all three settings. 
}
The best performance comes from deploying our MMPI for both modalities. The performance gain can be boiled down to the MMPI's ability to adaptively focus on the local regions of interest, as opposed to the naive cross-attention mechanism that attends to global features.

\begin{table*}
{
\caption{
{\color{blue}
Ablation on alternate interaction in the decoder. The experiments are conducted on the DeepInteraction++ framework.}
}
\label{tab:ablation_alternative}
\vspace{-3mm}
\begin{center}

\begin{tabular}{c|ccccccc}
\hline

\hline

\hline
  & NDS$\uparrow$ & mAP$\uparrow$ & mATE$\downarrow$ & mASE$\downarrow$ & mAOE$\downarrow$  & mAVE$\downarrow$  & mAAE$\downarrow$ \\
\hline

\hline
LiDAR-then-BEV & 72.8 & 70.4 & 27.0 & 25.3 & 29.3 & 23.7 & 18.7 \\
Alternate & \textbf{73.3} & \textbf{70.6} & \textbf{26.8} & \textbf{25.3} & \textbf{26.2} & \textbf{23.4} & \textbf{18.6} \\
\hline

\hline

\hline
\end{tabular}
\end{center}
}
\vspace{-3mm}
\end{table*}
\begin{table*}[htbp]
\caption{
Comparison with the LiDAR-only baseline Transfusion-L~\cite{bai2022transfusion} on nuScenes {\tt val} split. The mAP breakdown over categories is provided to demonstrate the improvement more comprehensively. ``C.V.'' and ``T.C.'' are abbreviations for ``construction vehicle'' and ``traffic cone''.
}
\renewcommand\tabcolsep{5.2pt}
\renewcommand\arraystretch{1.2}
\small
\begin{center}
 
\begin{tabular}{l|cc|cccccccccc}
\hline

\hline

\hline
{Method} & {mAP} & {NDS} & {Car} & {Truck} & {C.V.} & {Bus} & {Trailer} & {Barrier} & {Motorcycle} & {Bike} & {Pedestrain} & {T.C.} \\
\hline

\hline
Transfusion-L~\cite{bai2022transfusion} & 65.1 & 70.1 & 86.5 & 59.6 & 25.4 & 74.4 & 42.2 & 74.1 & 72.1 & 56.0 & 86.6 & 74.1 \\
Transfusion~\cite{bai2022transfusion} & 67.5
& 71.3 & 87.7 & 32.2 & 27.3 & 75.4 & 43.7 & 74.2 & 75.5 & 63.5 & 87.7 & 77.9 \\
\rowcolor[gray]{.9}
DeepInteraction  & 69.9 & 72.6 & {88.5} & {64.4} & {30.1} & {79.2} & {44.6}& {76.4} & {79.0}& {67.8} & {88.9} & {80.0}\\
\rowcolor[gray]{.9}
DeepInteraction++  & \textbf{70.6} & \textbf{73.3}  & \textbf{89.4} & \textbf{65.2} & \textbf{30.4} & \textbf{80.0} & \textbf{44.7} & \textbf{77.2} & \textbf{80.3} & \textbf{69.4} & \textbf{89.3} & \textbf{80.6} \\
\hline

\hline

\hline
\end{tabular}
\end{center}
\label{tab:nuscenes_val}
\vspace{-5mm}
\end{table*}

{\noindent \textbf{Alternate interaction.}}
{
\color{blue} 
The decoder introduced in Section~\ref{sec:decoder} alternately aggregates features from two modalities to maximize their utilization. To validate the effectiveness of this design, we compare it with a non-alternate design where the first three layers access only LiDAR features, followed by image features in the subsequent layers.
The results in Table~\ref{tab:ablation_alternative} show that our alternate interaction design yields a considerable advantage on mAOE and NDS, demonstrating its superiority in effectively aggregating multi-modal representation for decoding object attributes.
}

{\noindent \textbf{Number of decoder layers and queries.}}
As shown in Figure~\ref{fig:ablation_for_head_layer}, increasing the number of decoder layers up to 5 layers can consistently improve the performance for both models whilst introducing negligible latency.

Since our query embeddings are initialized in a non-parametric and input-dependent manner as in ~\cite{bai2022transfusion}, 
the number of queries is adjustable during inference. 
In Figure~\ref{tab:num_of_query}, we evaluate different combinations of query numbers used in training and testing on the DeepInteraction. 
Overall, the performance is stable over different choices with 200/300 for train/test as the best practice.

\subsubsection{Ablation on LiDAR backbones}
\begin{table}[ht]
\caption{
    Comparison for 3D detection with various point cloud backbones. 
}
\centering
\begin{tabular}{l|c|cc|cc}
\hline

\hline

\hline
\multirow{2}{*}{Methods} & \multirow{2}{*}{Modality} & \multicolumn{2}{c}{Voxel} & \multicolumn{2}{c}{Pillar} \\
 &  & mAP$\uparrow$ & NDS$\uparrow$ & mAP$\uparrow$ & NDS$\uparrow$ \\
\hline

\hline
PointPillars~\cite{pointpillars} & L & - & - & 46.2 & 59.1  \\
VoxelNet~\cite{pointpillars} & L & 52.6 & 63.0 & - & -  \\
Transfusion-L~\cite{bai2022transfusion} & L & 65.1 & 70.1 & 54.5 & 62.7 \\
Transfusion~\cite{bai2022transfusion} & L+C & 67.5 & 71.3 & 58.3 & 64.5 \\
\rowcolor[gray]{.9} 
DeepInteraction & L+C & 69.9 & 72.6 & 60.0 & 65.6 \\
\rowcolor[gray]{.9} 
DeepInteraction++ & L+C & \textbf{70.6} & \textbf{73.3} & \textbf{65.6} & \textbf{68.7} \\ 
\hline

\hline

\hline
\end{tabular}
\label{tab:backbone_pillar}

\end{table}

We examine the generalization ability of our framework with two different LiDAR backbones: PointPillars~\cite{pointpillars} and VoxelNet~\cite{zhou2018voxelnet}.
For PointPillars, we set the voxel size to (0.2m, 0.2m) while keeping the remaining settings as default.
For a fair comparison, we use the same number of queries as TransFusion~\cite{bai2022transfusion}.
As shown in Table~\ref{tab:backbone_pillar},
due to the proposed multi-modal interaction strategy, DeepInteraction exhibits consistent improvements over the LiDAR-only baseline using either backbone (by 5.5\% mAP for the voxel-based backbone, and 4.4\% mAP for the pillar-based backbone). 
These results manifest the generalization ability of our DeepInteraction across varying point cloud backbones. Critically, the improved interaction mechanism is
 particularly effective for the poor features extracted from light LiDAR backbone, exhibiting a stronger effect on the pillar backbone.

\subsubsection{Performance breakdown of each category}
To demonstrate more fine-grained performance analysis, we compare our DeepInteraction frameworks with the LiDAR-only baseline Transfusion~\cite{bai2022transfusion} at the category level in terms of mAP on nuScenes {\tt val} set. We can see from  Table~\ref{tab:nuscenes_val} that our fusion approach achieves remarkable improvements in all categories, 
especially in tiny or rare object categories.

\subsubsection{Component analysis of DeepInteraction++}
\begin{figure}
    \centering
    \includegraphics[width=1.0\linewidth]{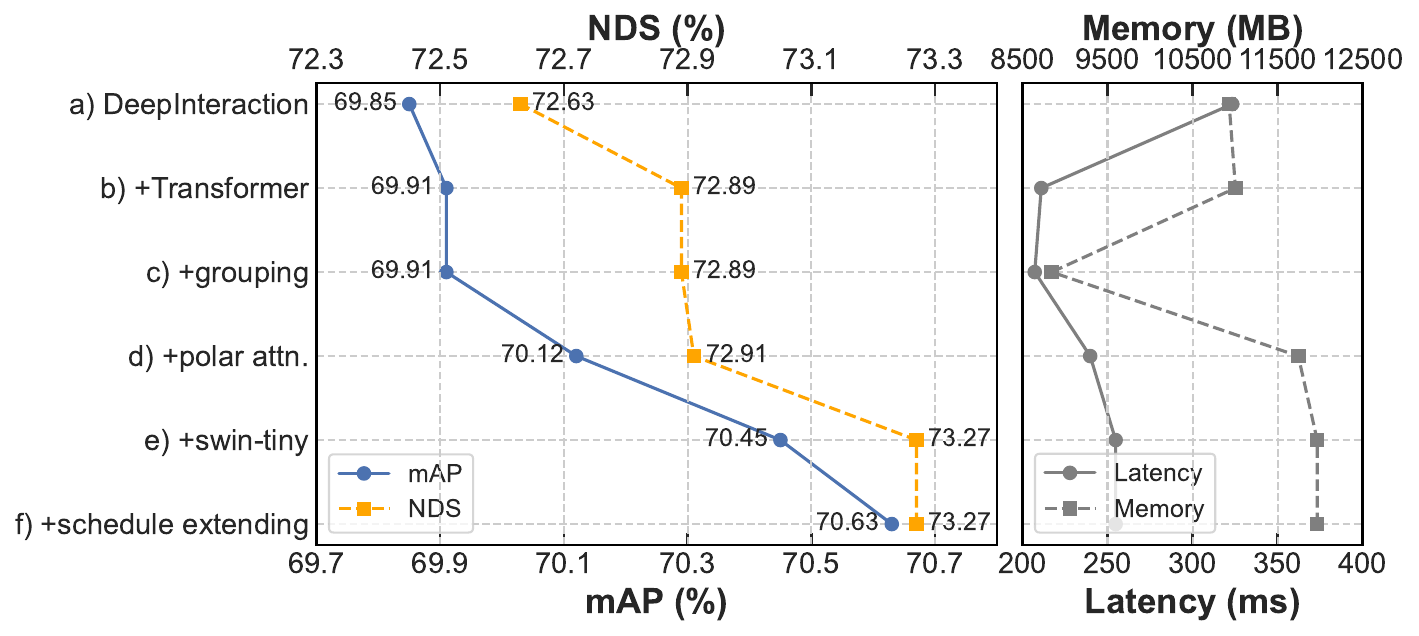}
    \caption{
        Improvement trajectory on 3D detection task. The latency is measured on NVIDIA RTX A6000 GPU.
    }
    \label{tab:journal_ablation}
\vspace{-5mm}
\end{figure}
In Figure~\ref{tab:journal_ablation}, we present the improvement from DeepIntection moving towards DeepInteraction++ step by step to demonstrate each design choice's effect and cost.

{\noindent \textbf{Transformer architecture with deformable attention.}}
In Section~\ref{sec:encoder}, we propose to instantiate representational interaction with a pair of parallel Transformers and replace the original stand-alone attention~\cite{ramachandran2019stand} used in IML with deformable attention~\cite{zhu2020deformable}. Comparing the a)-b) in Figure~\ref{tab:journal_ablation}, we can see that this modification effectively enhances both performance and efficiency. We consider that the performance gain may benefit from the more flexible receptive field of deformable attention, while the efficiency improvement is derived from the highly optimized Transformer implementation.

{\noindent \textbf{Grouped Image-to-LiDAR attention.}}
Although increasing the number of encoder layers can enhance performance, it comes at the cost of additional computation overhead. To compensate for these costs, we proposed grouped image-to-LiDAR attention in Section~\ref{sec:groupattn}. The results in Figure~\ref{tab:journal_ablation} c) demonstrate that introducing grouped attention significantly reduces memory usage without increasing latency thanks to the carefully designed grouping intervals.

{\noindent \textbf{LiDAR-guided cross plane polar attention.}
To further push performance, we introduce LiDAR-guided cross-plane polar attention for utilization of dense image features in Section~\ref{sec:polarattn}. The comparison between the c)-d) of Figure~\ref{tab:journal_ablation} validates the effectiveness of this mechanism. Introducing dense context information from image representation provides a beneficial complement to the original sparse interaction.

\begin{figure}[t]
    \centering
    \includegraphics[width=1\linewidth]{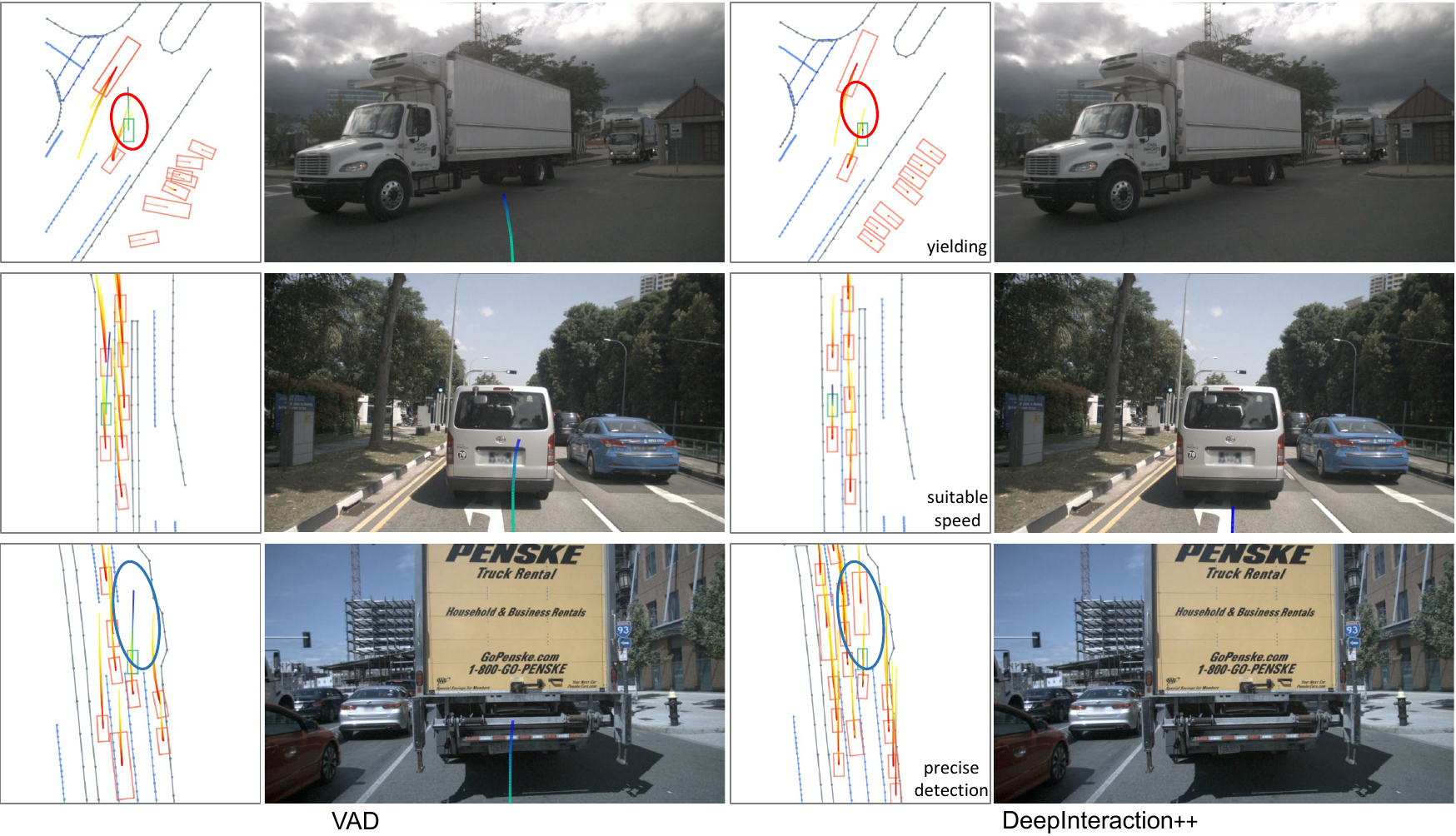}
    \caption{
    Qualitative comparison of end-to-end planning results between VAD~\cite{jiang2023vad} and our DeepInteraction++ on nuScenes {\tt val} set. In the HD map, the green box refers to the ego vehicle, and the circle parts highlight the significant differences.
    }
    \label{fig:plan_vis}
    \vspace{-5mm}
\end{figure}

{\noindent \textbf{Scaling backbone and training schedule.}}
In Figure~\ref{tab:journal_ablation} d)-e), we report the additional performance improvements brought by scaling the image backbones. It is worth noting that the improved representational interaction in DeepInteraction++ can further unleash the more powerful representation brought by the scaled backbone and achieve greater marginal gains than the original version as shown in Table~\ref{tab:journal_main}. Furthermore, the revised interaction structure mitigates the overfitting effect, allowing us to further push performance by extending the training schedule as shown in Figure~\ref{tab:journal_ablation} f).

\begin{figure}[t]
    \centering 
    \setlength{\tabcolsep}{2.0pt}
    \begin{tabular}{c}
    \includegraphics[width=1\linewidth,clip]{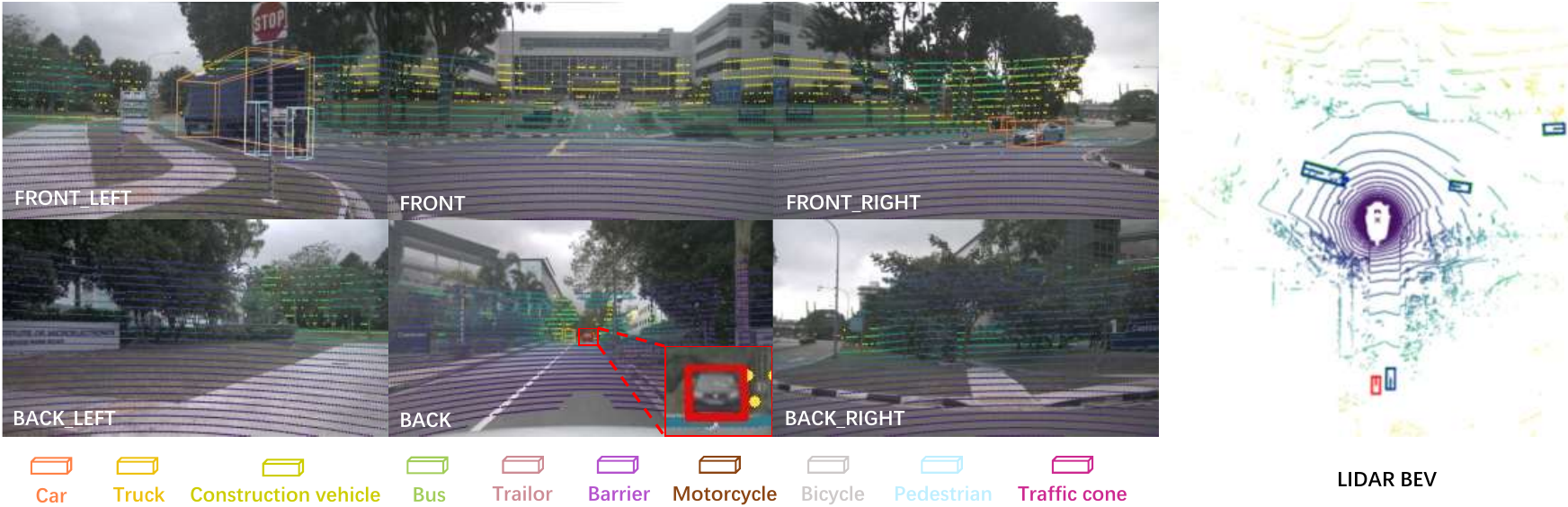}
\\
    \includegraphics[width=1\linewidth,clip]{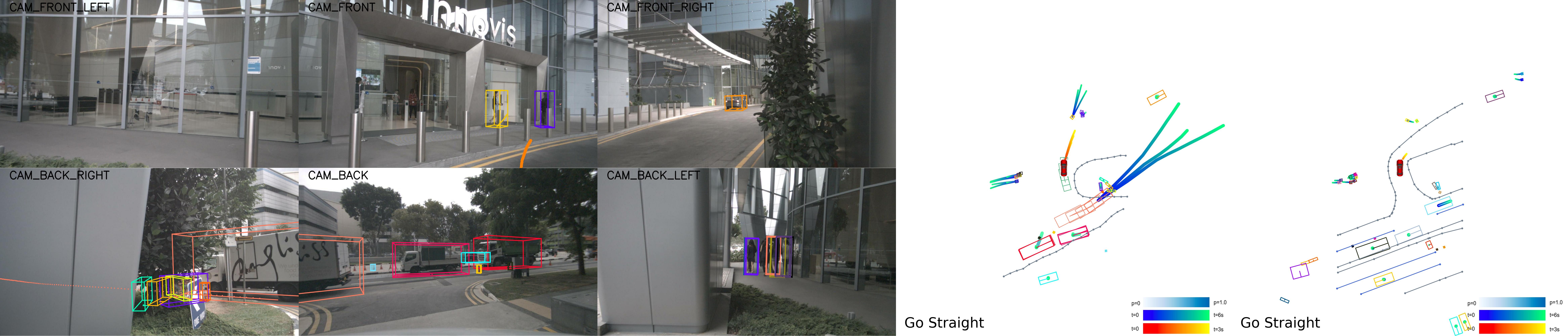}
    \end{tabular}
\caption{
{
\color{blue} Failure cases of detection (top) and planning (bottom). The missed ground truth box is highlighted in red.}
}
    \label{fig:failure_case}
\end{figure}
\begin{table*}
\caption{
Comparison of end-to-end planning performance with state-of-the-art methods on the nuScenes {\tt val} set.
}
\renewcommand\tabcolsep{7pt}
\renewcommand\arraystretch{1.2}
\small
\centering

\begin{tabular}{l|c|cccc|cccc}
\hline

\hline

\hline
\multirow{2}*{Method}& \multirow{2}*{Present at} & \multicolumn{4}{c}{L2(m)$\downarrow$}\vline & \multicolumn{4}{c}{Col. Rate ($\%$)$\downarrow$} \\ 
 & & 1s & 2s & 3s & Avg. & 1s & 2s & 3s & Avg. \\
\hline

\hline
NMP~\cite{zeng2019end}& CVPR'19 & - & - & 2.31 & - & - & - & 1.92 & - \\
SA-NMP~\cite{zeng2019end} & CVPR'19 & - & - & 2.05 & - & - & - & 1.59 & - \\
FF~\cite{hu2021safe} & CVPR'21 & 0.55 & 1.20 & 2.54 & 1.43 & 0.66 & 0.17 & 1.07 & 0.43 \\
EO~\cite{khurana2022differentiable} & ECCV'22 & 0.67 & 1.36 & 2.78 & 1.60 & 0.04 & 0.09 & 0.88 & 0.33 \\
ST-P3~\cite{hu2022st} & ECCV'22 & 1.33 & 2.11 & 2.90 & 2.11 & 0.23 & 0.62 & 1.27 & 0.71 \\
UniAD~\cite{hu2023planning} & CVPR'23 & 0.48 & 0.96 & 1.65 & 1.03 & \textbf{0.05} & 0.17 & 0.71 & 0.31 \\
GPT-Driver~\cite{mao2023gpt} & arXiv'23 & \textbf{0.27} & 0.74 & 1.52 & 0.84 & 0.07 &0.15& 1.10 & 0.44\\
VAD\cite{jiang2023vad}  & ICCV'23 & 0.41 & 0.70 & \textbf{1.05} & 0.72 & 0.07 & 0.17 & 0.41 & 0.22 \\
\rowcolor[gray]{.9} 
DeepInteraction & NeurIPS'22& 0.40 & 0.71 & 1.13  & 0.75 & 0.07 & 0.17 & 0.52 & 0.25 \\
\rowcolor[gray]{.9} 
\textbf{DeepInteraction++} & Submission & 0.36 & \textbf{0.67} & 1.06  & \textbf{0.70} & \textbf{0.05} & \textbf{0.15} & \textbf{0.38} &\textbf{0.19} \\
\hline

\hline

\hline
\end{tabular}
\label{tab:plan}
\end{table*}
\begin{table}[t]
\caption{
Comparison of the perception and prediction results with VAD-base on the nuScenes {\tt val} set.
}
\renewcommand\tabcolsep{4pt}
\centering

\begin{tabular}{l|cc|ccc}
\hline

\hline

\hline
\multirow{2}*{Method} & \multicolumn{2}{c}{Detection}\vline & \multicolumn{3}{c}{Prediction} \\ 
& mAP$\uparrow$ & NDS$\uparrow$ & minADE$\downarrow$ & minFDE$\downarrow$ & MR$\downarrow$ \\
\hline

\hline
UniAD~\cite{hu2023planning} & - & - & 0.728 & 1.054 & 0.154 \\
VAD~\cite{zeng2019end} & 0.330 & 0.460 & 0.682 & 0.881 & 0.083 \\
\rowcolor[gray]{.9} 
DeepInteraction & 0.492 & 0.613 & 0.445 & 0.689 & 0.072 \\
\rowcolor[gray]{.9} 
\textbf{DeepInteraction++} & \textbf{0.557} & \textbf{0.660} & \textbf{0.337} & \textbf{0.539} & \textbf{0.047} \\
\hline

\hline

\hline
\end{tabular}
\label{tab:e2eper}
\vspace{-5mm}
\end{table}

\subsection{Extension to the end-to-end planning}

\paragraph{Experimental setup} 
We train the e2e framework with the same settings as the detection task, except for a batch size of 1. As for metrics, $minADE$, $minFDE$, and $MR$ across six prediction modes are employed for the evaluation of prediction performance, while ego-trajectory displacement error (L2) and Collision Rate are adopted in the planning task.

\paragraph{Performance comparison and qualitative analysis} 
Benefiting from the multi-modal representation and interactive decoding, our e2e extension of DeepInteraction++ achieves better perception and prediction performance compared to VAD~\cite{jiang2023vad}, as shown in Table~\ref{tab:e2eper}. Moreover, we report the planning results in Table~\ref{tab:plan}, which demonstrates that DeepInteraction++ remarkably surpasses existing planning-oriented methods on most evaluation metrics. Besides providing a more accurate planning trajectory, DeepInteraction++ can achieve a lower collision rate by resorting to more precise and comprehensive perception and prediction for traffic participants. Furthermore, we also implement an end-to-end framework based on the original DeepInteraction, which takes the sparse points as a medium for representation interaction. In comparison, the DeepInteraction++ can preserve more road elements from images thorough deformable attention and dense polar interaction, achieving superior performance across all metrics.

To intuitively demonstrate the superiority of DeepInteraction++, we provide several qualitative results in Figure~\ref{fig:plan_vis}. By integrating multi-modal information and employing a meaningful fusing strategy, our method can comprehensively understand and analyze the driving scenario, hence giving more reasonable planning action even in a complex and intricate driving environment. For example, the yielding action and suitable speed are adopted in the first two cases. Besides, due to the precise upstream perception, DeepInteraction++ is able to effectively avoid the incorrect actions caused by cumulative error as shown in the third row. 

\subsection{Failure cases and discussions}
{
\color{blue}
In Figure~\ref{fig:failure_case}, we present several failure cases to provide a more comprehensive perspective on the limitations of the proposed framework and shed light on potential challenges may face in practice. 
For scene perception, our explicit LiDAR-guided 3D mapping makes the model susceptible to misalignment or sensor failures. For instance, if an object lacks LiDAR signals, it may be missed by the detector. While the integration of learning-based polar ray attention helps mitigate this issue to some extent, it still occurs in certain cases, as illustrated in the top plot of Figure~\ref{fig:failure_case}.
For planning tasks, although multi-sensor fusion provides richer scene information, challenges such as map segmentation and motion prediction remain not fully resolved within existing frameworks. As a result, this can lead to unreasonable planning trajectories, as shown in the bottom plot of Figure~\ref{fig:failure_case}.
}

\section{Conclusion}
In this work, we have presented a novel multi-modality interaction approach 
for exploring both the intrinsic multi-modal complementary nature and their respective characteristics in autonomous driving.
This key idea is to maintain two modality-specific representations and establish interactions between them for both representation learning and predictive decoding.
This strategy is designed particularly to resolve the fundamental limitation of existing unilateral fusion approaches that image representation is insufficiently exploited due to their auxiliary-source role treatment.
Extensive experiments demonstrate our approach yields state-of-the-art performances on the highly-competitive nuScenes benchmark,
across both 3D object detection and end-to-end autonomous driving tasks.

\section*{Acknowledgments}
This work was supported by the National Natural Science Foundation of China (Grant No. 62376060).

\bibliographystyle{IEEEtran}
\bibliography{ref}

\end{document}